\documentclass[final,3p,times]{article}

\usepackage{amssymb}
\usepackage{amsmath} 
\usepackage{PRIMEarxiv}

\usepackage{latexsym}
\usepackage[latin1]{inputenc}
\usepackage{times}
\usepackage[linesnumbered,ruled,lined,boxed]{algorithm2e}
\usepackage{algorithmic}

\usepackage{epsfig,shadow,amssymb,bm}
\usepackage{booktabs}
\usepackage{multirow}
\usepackage{graphicx}
\usepackage{tikz}
\usepackage{stackrel}
\usepackage{pdflscape}[2016/05/14]
\usepackage{rotating}
\usepackage{tablefootnote}
\usepackage{dsfont}
\usepackage{hyphenat}

\usepackage{longtable}
\usepackage{enumitem}

\usepackage{colortbl}
\usepackage{xurl}
\usepackage{subcaption}
\usepackage[breaklinks=true]{hyperref}

\usetikzlibrary{shapes,arrows}
\usetikzlibrary{positioning}
\usetikzlibrary{arrows,automata,decorations,shapes}
\tikzset{
	treenode/.style = {shape=rectangle, rounded corners,
		draw, align=center,
		top color=white, bottom color=blue!20},
	root/.style     = {treenode, font=\Large, bottom color=red!30},
	env/.style      = {treenode, font=\ttfamily\normalsize},
	dummy/.style    = {circle,draw}
}

\newcommand{\CASE}[1]{\STATE \textbf{case} #1\textbf{:} \begin{ALC@g}}
	\newcommand{\ENDCASE}{\end{ALC@g}}

\newcommand{\DEFAULT}{\STATE \textbf{default:} \begin{ALC@g}}
	\newcommand{\ENDDEFAULT}{\end{ALC@g}}
\newcommand{\DEFAULTLINE}[1]{\STATE \textbf{default:} }

\title{Embedded feature selection in LSTM networks with multi-objective evolutionary ensemble learning for time series forecasting}

\author{Raquel Espinosa, Fernando Jim\'enez, Jos\'e Palma\\
	Artificial and Knowledge Engineering Group\\
	University of Murcia\\ 
	Spain \\
	\texttt{\{raquel.espinosa,fernan,jtpalma\}@um.es} \\
}

\begin{document}
\maketitle

\begin{abstract}
Time series forecasting plays a crucial role in diverse fields, necessitating the development of robust models that can effectively handle complex temporal patterns. In this article, we present a novel feature selection method embedded in Long Short-Term Memory  networks, leveraging a multi-objective evolutionary algorithm. Our approach optimizes the weights and biases of the LSTM in a partitioned manner, with each objective function of the evolutionary algorithm targeting the root mean square error  in a specific data partition. The set of non-dominated forecast models identified by the algorithm is then utilized to construct a meta-model through stacking-based ensemble learning. Furthermore, our proposed method provides an avenue for attribute importance determination, as the frequency of selection for each attribute in the set of non-dominated forecasting models reflects their significance. This attribute importance insight adds an interpretable dimension to the forecasting process. Experimental evaluations on air quality time series data from Italy and southeast Spain demonstrate that our method substantially improves the generalization ability of conventional LSTMs, effectively reducing overfitting. Comparative analyses against state-of-the-art CancelOut and EAR-FS methods highlight the superior performance of our approach.
\end{abstract}



\section{Introduction} 
\textit{Feature selection} (FS) \cite{Butcher2020} is a crucial step in machine learning that involves selecting a subset of relevant features from the original set of input variables (features) to improve model performance and interpretability. By eliminating irrelevant or redundant features, FS helps reduce overfitting, enhance generalization, and speed up the learning process.

FS methods can be \textit{attribute evaluation methods} or \textit{attribute subset evaluation methods}. The first evaluate each attribute independently, and can be \textit{univariate} or \textit{multivariate}. These methods calculate a ranking score or measure for each feature, so they are also called \textit{feature ranking} methods. The second evaluate subsets of attributes in batches, and are multivariate methods since they always consider interactions between attributes. On the other hand, FS methods can also be classified into \textit{filter}, \textit{wrapper} and \textit{embedded} methods. Filter methods assess the relevance of attributes or attributes subsets independently from any learning algorithm. They rely on statistical measures, such as correlation, consistency, redundancy, and many others. Wrapper methods evaluate attributes or attributes subsets using a specific learning algorithm. Wrapper methods can be very accurate but can also be computationally expensive. Finally, embedded methods incorporate feature selection as part of the model training process, offering a balance between efficiency and feature interactions. The choice of method depends on factors such as the dataset size, feature space complexity, computational resources, and the desired trade-off between accuracy and computational cost.

Another topic of the present work is \textit{long short-term memory} (LSTM) networks \cite{10.1162/neco.1997.9.8.1735}. LSTMs are a specialized type of \textit{recurrent neural network} (RNN) architecture. They were designed to overcome the limitations of traditional RNNs when dealing with long-term sequential data, such as time series. LSTMs are built on the concept of long-term and short-term memory. They have an intricate internal structure that enables them to remember and learn long-term dependencies in the input data. Unlike standard RNNs, LSTMs incorporate memory units called \textit{cells} and \textit{gates} that regulate the flow of information within the network.
Key features of LSTMs include:

\begin{itemize}
\item[--] Long-term memory: LSTMs can learn to retain relevant information over long periods, avoiding the ``vanishing gradient'' problem that affects traditional RNNs.

\item[--] Gate mechanisms: LSTMs have three main gates: the forget gate, the input gate, and the output gate. These gates control the flow of information within the network, allowing relevant information to be retained or forgotten.

\item[--] Flexible sequence length: LSTMs are not constrained by the length of input sequences. They can process and learn from long sequences of data while maintaining the ability to capture long-term temporal dependencies.
\end{itemize}

LSTMs find applications in various fields, including \textit{natural language processing} \cite{10.1145/3544109.3544157}, \textit{speech recognition} \cite{9733937}, and \textit{time series forecasting} \cite{ESPINOSA2021107850}. LSTMs are particularly suitable for time series forecasting due to their ability to capture long-term dependencies in data. They are used in financial forecasting, weather prediction, sales analysis, air quality, and any context where modeling and predicting patterns in time sequences is required. The suitability of LSTMs for time series forecasting lies in their capacity to capture long-term temporal dependencies, enabling the network to learn complex patterns in the data. This is beneficial for analyzing and forecasting time series, as LSTMs can retain relevant historical information and utilize it to make accurate predictions about future values in the series.

LSTMs are capable of automatically learning relevant features from the input data, extracting high-level representations, and making predictions based on the learned patterns.  However, filter FS techniques can be applied as a preprocessing step before feeding the data into an LSTM model \cite{Esp2023a}. By selecting the most informative features, we can improve the LSTM's performance by providing it with a reduced and more relevant feature set. Other FS methods used in conjunction with LSTMs are the \textit{surrogate methods} \cite{Esp2023b}, which consist in the use of metamodels to evaluate the subsets of attributes, allowing to reduce the computational time required by conventional FS wrapper methods. Finally, the embedded FS methods can also be used with LSTMs. One common approach for embedded feature selection in LSTMs is through the use of \textit{attention} mechanisms \cite{Katrompas2022}. Attention mechanisms allow the LSTM model to dynamically focus on different parts of the input sequence, assigning different weights or importance to different time steps or features. By doing so, the model can effectively attend to the most informative features at each time step, while downplaying the influence of less relevant features.

Usually, gradient descent techniques are used to adjust the weights and biases of the LSTM network. However, using gradient descent techniques to train LSTM networks can indeed lead to local optima. The problem of local optima occurs when the gradient descent algorithm gets trapped in a local minimum of the loss function instead of reaching the desired global minimum. This can happen because local gradients lead to a local minimum but not the global minimum, resulting in suboptimal performance of the LSTM network. To mitigate the issue of local optima, various techniques have been proposed. Some of these techniques include proper weight initialization, adaptive learning rate schemes, exploring different network architectures, and using more advanced optimization algorithms such as \textit{stochastic gradient descent with momentum} \cite{10.5555/3495724.3497257}, \textit{second-order optimization methods} like the \textit{Newton method} \cite{Tan2019}, or global optimization methods like \textit{evolutionary algorithms} \cite{10.1145/3520304}.

Finally, overfitting is a common problem that can occur when training LSTM networks, as well as other types of neural networks. It refers to a situation where the network becomes too specialized in the training data and fails to generalize well to unseen data. To address the overfitting problem in LSTM networks, several techniques can be employed, e.g. data augmentation, early stopping, model complexity control, or \textit{regularization} techniques like \textit{L1} or \textit{L2}, \textit{dropout}, or \textit{batch normalization}.

In this context, we propose a novel embedded FS method based on LSTM networks, called EFS-LSTM-MOEA (\textit{Embedded Feature Selection in LSTM networks with Multi-Objective Evolutionary Algorithms}). We use global optimization techniques, specifically \textit{multi-objective evolutionary algorithms} (MOEA) \cite{DBLP:conf/emo/2023}, to adjust the weights and biases of the LSTM network. The proposed method has been applied to air quality problems in south-eastern Spain and Italy where the authors develop research projects and has been compared with conventional LSTM and with two state-of-the-art FS methods for RNNs. The main contributions of the work are the following:

\begin{enumerate}
\item In order to prevent overfitting, the training set is split into $ n $ training subsets which are associated, each of them, to an objective function of a multi-objective optimization problem that is solved with a MOEA. The set of non-dominated solutions found by the MOEA are a set of LSTM networks that are finally assembled to build a forecast model using ensemble learning, particularly with the \textit{stacking} technique. In this way, the training dataset is distributed among the multiple LSTM networks, thus increasing the generalization capacity of the forecast model.
The authors of the present work proposed in \cite{ESPINOSA202215} a similar technique based on MOEAs. However, 1) the purpose of the MOEA in \cite{ESPINOSA202215} was for spatio-temporal forecasting with time series data from different monitoring stations, while in the present work the purpose of the MOEA is overfitting prevention, 2) in \cite{ESPINOSA202215} the MOEA searches for forecast models based on multiple linear regression, while here the MOEA searches for LSTM networks, and 3) in \cite{ESPINOSA202215} embedded FS is not performed as in the present work.

\item The proposed MOEA optimizes the weights and biases of the LSTM and performs FS simultaneously. For this, the individuals of the MOEA are represented with mixed encoding: binary to encode the selected attributes, and floating point to encode the weights and biases of the LSTM. 

\item We propose a new measure of \textit{feature importance} based on the frequency of occurrence of the feature in the set of LSTM networks found by the MOEA.
\end{enumerate}

The rest of the manuscript has been organized as follows: Section \ref{RelatedWorks} describes the most important related works of the state of the art; Section \ref{framework} describes the main components of the proposed surrogate-assisted MOEA for FS; Section \ref{ER} shows the performed set of experiments; Section \ref{AnaResDis} analyses and discusses the results; finally, Section \ref{Conclusion} draws the main conclusions of the research and points out future work.

\section{Related works} 
\label{RelatedWorks}

This section presents state-of-the-art work related to FS  in neural networks, with special emphasis on embedded FS methods in LSTM networks and methods based on evolutionary search. 


Among the works on FS embedded in neural networks we can find that of Varma \cite{k2020embedded}, which proposes two different approaches \textit{sensitivity based selection} (SBS) and \textit{stepwise weight pruning algorithm} (SWPA). On the one hand, SBS computes the gradients of the output label with respect to the input features, and quantifies the importance of each feature. On the other hand, SWPA sequentially removes the weights from the network, aiming to minimize the loss in the model performance while reducing the number of parameters. Wei \textit{et al.} present  \cite{10.1145/3534678.3539290} a novel framework called \textit{dual-world embedded attentive FS} (D-AFS) for reducing industrial sensor requirements. D-AFS incorporates both the real world and a virtual peer with distorted features. By analyzing the deep reinforcement learning algorithms response in these two worlds, D-AFS can quantitatively identify the importance of different features for control. Perin \textit{et al.} \cite{PerinWuPicek2022} paper focuses on the role of FS in deep learning-based profiling attacks for side-channel analysis. The experiments utilize \textit{multi-layer perceptrons} (MLPs) and CNNs, with up to eight hidden layers, in three FS scenarios (from minimal features to using all features).  Yang \textit{et al.} \cite{Yang2010} propose wrapper-based FS method specifically designed for MLP neural networks. The method utilizes a feature ranking criterion that measures the importance of each feature by comparing the probabilistic outputs of the MLP with and without the feature across the feature space.  Atashgahi \textit{et al.} \cite{2023arXiv230307200A} introduce a resource-efficient approach for supervised FS using sparse neural networks, called \textit{NeuroFS}. By gradually pruning uninformative features from the input layer of a sparse neural network trained from scratch, NeuroFS efficiently derives an informative subset of features. Cancela \textit{et al.} \cite{9983480} develop a new embedded FS algorithm called \textit{End-to-End Feature Selection} that combines accuracy and explainability. By utilizing gradient descent techniques, the algorithm incorporates non-convex regularization terms to enforce the selection of a maximum number of features for subsequent use by the classifier. A novel heart disease prediction model is proposed by Zhang \textit{et al.} \cite{zhang2021heart}, combining an embedded FS method and deep neural networks. The FS method employs the \textit{LinearSVC} algorithm with L1 norm penalty to select a subset of features significantly associated with heart disease. These selected features are then fed into a deep neural network with weights initialized using the \textit{He} initializer. Wright \textit{et al.} \cite{Wright2011} introduce a FS algorithm called \textit{incremental feature selection embedded in NEAT} (IFSE-NEAT) is introduced, which integrates sequential forward search into the neuroevolutionary algorithm \textit{neuroevolution of augmenting topologies} (NEAT). The main objective of IFSE-NEAT is to identify the most relevant features for effective policy learning in real-world environments.  Mirzaei \textit{et al.} \cite{mirzaei2019deep} propose a \textit{teacher-student feature selection} (TSFS) method to address the challenges of high-dimensional data in machine learning. The TSFS method utilizes a `teacher' network, such as a deep neural network or dimension reduction method, to learn a low-dimensional representation of the data. Then, a `student' network, a simpler neural network, performs FS by minimizing the reconstruction error of the low-dimensional representation. 
Gu \textit{et al.} \cite{gu2021feature} address the challenge of high dimensionality in multivariate time series  data by proposing a novel neural component called \textit{neural feature selector} (NFS). NFS consists of two modules: temporal \textit{convolutional neural networks} (CNNs) process each feature stream independently, and an aggregating CNN combines the processed streams for downstream networks.  Figueroa Barraza \textit{et al.} \cite{s21175888} propose a technique for FS embedded in deep neural networks to address the tradeoff between accuracy and interpretability in prognostics and health management models. The technique involves a FS layer that evaluates the importance of input features. A new metric called \textit{ranking quality score} is introduced to measure the performance evolution based on feature rankings. Borisov \textit{et al.} \cite{cancelout} present a layer called \textit{CancelOut} for deep neural networks to perform feature ranking and FS tasks in both supervised and unsupervised learning. 
Liu \textit{et al.} \cite{LIU2022109715} employ a Gaussian mixture model in a CNN to approximate the feature distribution within each subclass in a image classification problem. 
Liu \textit{et al.} \cite{liu2021using} propose a hybrid lightweight intrusion detection system for cyberattacks that combines an embedded model for FS and a CNN for attack detection and classification. This system has two models one based on \textit{random forest} (RF) and the other based on \textit{extreme gradient boosting}, both combined with a CNN. Bo \textit{et al.} \cite{Bo2006} introduce a neural network model for cancer classification called \textit{multi-layer perceptrons with embedded feature selection} (MLPs-EFS), which incorporates FS into the training process. 

Particularly, several embedded FS methods are based on or tested with LSTM networks, as is the case with Pai and Ilango \cite{9243376} who employ for financial time series forecasting different techniques such as recursive feature elimination, correlation, and random forest algorithms to determine feature importance, and the results are evaluated using LSTM networks. Chowdhury \textit{et al.} \cite{9006432} present a novel framework based on RNN for FS in micro-array data analysis. The framework involves dividing the features into selected and candidate categories, and iteratively selecting or excluding features based on gradient computations and backpropagation. Different architectures of recurrent models, including GRU, LSTM, and bi-directional LSTM, are implemented within the framework.
\c{S}ahín and D\'{\i}r\'{\i} \cite{ahn2019RobustFS} propose an ensemble gene selection framework for sequence modelling applied to biomarker discovery. The proposed method consists of creating a model to learn long sequences with an LSTM embedded into an \textit{artificial immune recognition system} for robust FS.

Attention-based FS has emerged as a promising approach to effectively address the challenges posed by high-dimensional data as demonstrated by several studies. As in the case of Gui \textit{et al.} \cite{gui2019afs}, which introduce \textit{attention-based feature selection} (AFS), a novel neural network-based architecture designed to address the challenges of FS in high-dimensional and noisy datasets. AFS incorporates an attention module to generate feature weights by formulating the problem as a binary classification task for each feature. Yasuda \textit{et al.} \cite{yasuda2023sequential} proposes \textit{sequential attention}, an adaptive FS algorithm based on attention for neural networks. Sequential Attention considers the residual value of features during selection and utilizes attention weights as a measure of feature importance. Wang \textit{et al.} \cite{WANG2023129200} present a feature attention mechanism model for the extraction of effective hydrological correlation features and accurate runoff prediction. The model transforms the FS problem into binary classification problems and assigns attention units to specific features in the network, updating feature weightings through a learning module. Cao \textit{et al.} \cite{cao2021multiattention} introduces a \textit{multiattention-based supervised FS} method for multivariate time series prediction. The input data is sliced into 1D from two orthogonal directions, and each attention module generates weights from their respective dimensions. A global weight generation method is proposed to facilitate FS.  Xue \textit{et al.} \cite{XUE2023111084} 
propose a \textit{feature ranking} method (EAR-FS) in which the importance of features is pre-learned using an MLP with an embedded attention mechanism. Su \textit{et al.} \cite{su2023modular} propose an interpretable neural network model that incorporates an attention-based feature selection component to eliminate irrelevant features from the temporal dependencies learned by an RNN. The selected features are used independently to provide the user with interpretability of the model.

There are some studies that have combined attention-based mechanisms with LSTMs applied to FS, such as Wu \textit{et al.} \cite{10.1117/12.2680970} which introduce a novel load forecasting model utilizing double attention LSTM and FS. By integrating the LSTM with dual attention mechanisms, the model dynamically explores the relationship between load and input features, leading to improved forecasting accuracy. Shen \textit{et al.} \cite{10.1145/3569422} focus on power forecasting in data centers using FS based on causality to identify the influential metrics. The model combines LSTM, tensor techniques, and attention mechanisms to capture temporal patterns effectively. Nie \textit{et al.} \cite{2023} propose a method based on multi-head attention to accurately predict the remaining useful life of degrading equipment. The method incorporates an adaptive FS technique to identify the optimal subset of features. The selected features are then input into a \textit{multi-head temporal convolution network-bidirectional LSTM} network, where each feature is individually mined to maximize information integrity. Chen \textit{et al.} \cite{Chen2023} propose a fusion architecture for dynamic gesture recognition that combines convnets-based models and LSTM-based models in parallel. The model utilizes attention mechanisms, such as channel attention and spatial attention, to extract relevant features and adjust network contributions. Zhang \textit{et al.} \cite{Zhang2019} present an \textit{attention-based LSTM} model for financial time series prediction. The proposed framework consists of two stages: an attention model assigns weights to input features, and the attention feature is used to select relevant feature sequences for prediction in the next time frame using an LSTM neural network.

Another related topic is the evolutionary learning of the neural network, in particular, the evolutionary search for the weights and biases of the neural network units. Li \textit{et al.} \cite{DBLP:journals/corr/abs-1811-03760} propose an attention-based LSTM for time series prediction where attention layer weights are optimized by evolutionary computation. Santra and Lin \cite{en12112040} propose a genetic algorithm where each individual is used to initialize the weight matrix of an LSTM, and the fitness of each individual is the \textit{mean absolute percentage error} of the LSTM in the training set. It is important to note that the genetic algorithm does not optimize the final weights of the LSTM, but instead searches for the optimal weight matrix to initialize the LSTM. Baioletti \textit{et al.} \cite{math8010069} propose \textit{differential evolution} with \textit{rand/1} mutation to optimize the weights and bias of a \textit{feedforward neural network} composed of $L$ layers. To reduce the computational time, the authors implement a batching method to compute the fitness of the individuals with which the training dataset is divided into $ k $ batches of the same size. The population is then tested on a set of training samples instead of the entire training set. This set of training samples constitutes a window $ U $ that contains samples from a batch and is updated every certain number of generations with samples from the next batch to make a smooth transition between batches. Saito \textit{et al.} \cite{Saito2018} present a novel approach for embedded FS using probabilistic model-based evolutionary optimization. The FS process involves utilizing the multivariate Bernoulli distribution to determine the optimal features. During the training phase, the distribution parameters are optimized through an update rule inspired by \textit{population-based incremental learning}. Simultaneously, a neural network, undergoes parameter updates using gradient descent.

\subsection{Conclusions}
To the best of our knowledge, it is the first time that the weights and biases of an LSTM network are fully optimized with an evolutionary algorithm. In addition to being a gradient-free global search process, the evolutionary algorithm has allowed embedding feature selection in the representation of individuals, and perform multi-objective optimization with multiple partitions of the data with the purpose of increasing the generalization capacity of the final model. Due to all these characteristics, we think that the method proposed in this paper is a novel alternative that can compete with the state-of-the-art embedded FS methods for LSTM networks, as we demonstrate with the experiments carried out.

\section{A multi-objective evolutionary ensemble learning approach for embedded feature selection in LSTM networks for time series forecasting}
\label{framework}

In this section the proposed method is described. Section \ref{GM} presents the general framework of the method. Section \ref{MOEA} describes the main components of the proposed MOEA. Section \ref{Stacking} presents the stacking-based ensemble learning algorithm for the construction of the forecast meta-model from the set of non-dominated LSTM models found by the MOEA. Section \ref{Multistep} shows the multi-step forecasting strategy used to generate the $h$-step ahead predictions from the built forecast model. Finally, Section \ref{FI} describes the proposed method to obtain the importance of attributes.

\subsection{General framework}
\label{GM}
\begin{figure*}[tp]
	\centering
	\resizebox{0.8\textwidth}{!}{\includegraphics[width=\textwidth]{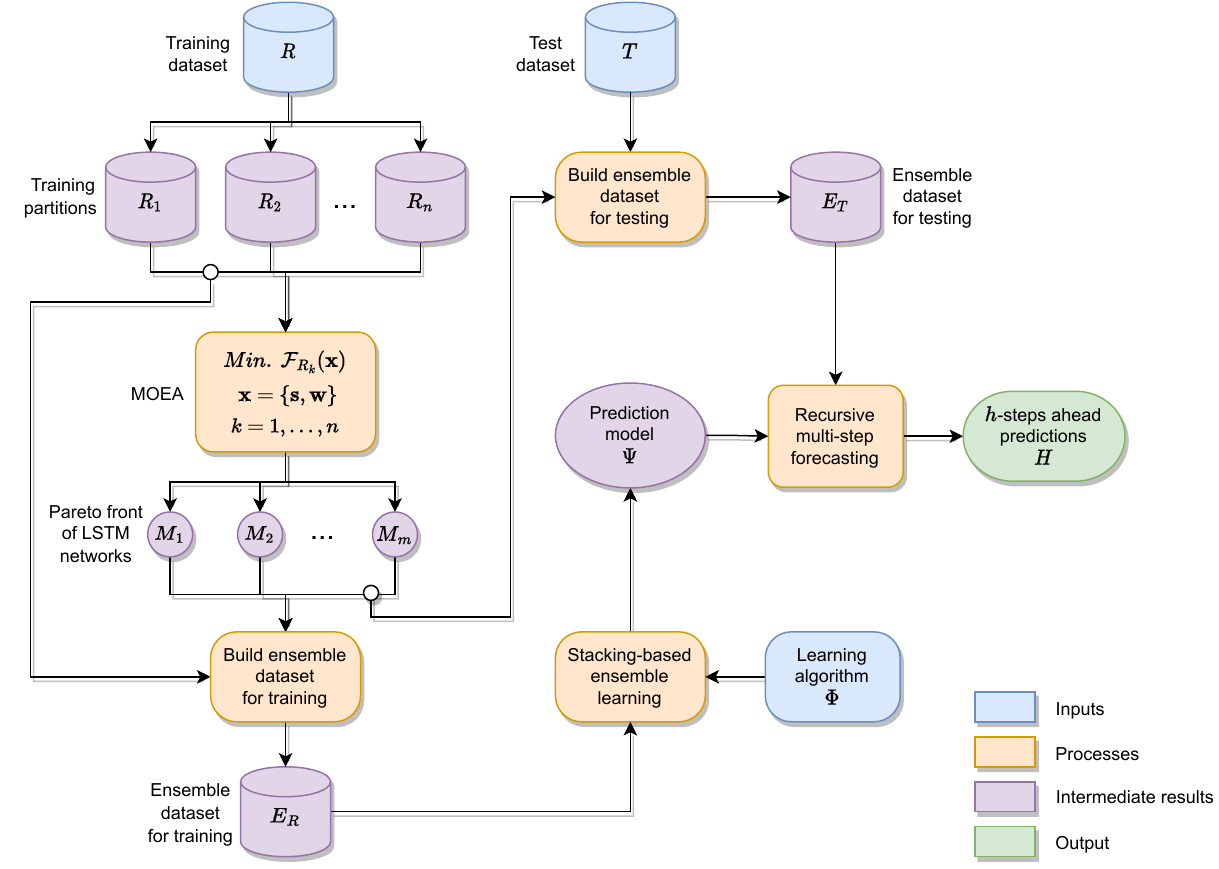}}
	\caption{General framework of EFS-LSTM-MOEA.}
	\label{fig:methodology}
\end{figure*}

Next, we describe the general framework of EFS-LSTM-MOEA, graphically illustrated as a flowchart in Figure \ref{fig:methodology}, and as pseudo-code in Algorithm \ref{alg:EFS-LSTM-MOEA}.

\begin{algorithm}[!ht]
	\caption{EFS-LSTM-MOEA}
	\label{alg:EFS-LSTM-MOEA}
	\begin{algorithmic}[1]
		\REQUIRE $R$ \COMMENT{Training dataset}
		\REQUIRE $T$ \COMMENT{Test dataset}
		\REQUIRE $n$ \COMMENT{Number of training partitions}
		\REQUIRE $P>1$ \COMMENT{Population size}
		\REQUIRE $G>1$ \COMMENT{Number of generations}
		\REQUIRE $0\leq p_c\leq 1$ \COMMENT{Crossover probability}
		\REQUIRE $0\leq p_m\leq 1$ \COMMENT{Mutation probability}
		\REQUIRE $\Phi$ \COMMENT{Algorithm for stacking-based ensemble learning}
		\ENSURE $H$ \COMMENT{$h$-steps ahead predictions for test dataset $T$}
		\STATE $R_1,\ldots,R_n\leftarrow$ Split($R,n$)
		\STATE $M_1,\ldots,M_m\leftarrow$ MOEA($R_1,\ldots,R_n,P,G,p_c,p_m$) 
		\STATE $E_R\leftarrow$ EnsembleTrainingDataset($R_1,\ldots,R_n,M_1,\ldots,M_m$)
		\STATE $\Psi\leftarrow$ EnsembleLearning($\Phi,E_R$)
		\STATE $E_T\leftarrow$ EnsembleTestDataset($T,M_1,\ldots,M_m$)
		\STATE $H\leftarrow$ RecursiveMultiStepForecasting($\Psi,E_T$)
		
		\RETURN {$H$} 
	\end{algorithmic}
\end{algorithm}

We consider a normalized time series dataset $D=\big\{\{\textbf{d}_1,o_1\},\ldots,\{\textbf{d}_{r},o_r\}\big\}$  with $r$ samples. We assume that $D$ has been previously transformed with the sliding window method \cite{brownlee2020}. Each sample $\{\textbf{d}_t,o_t\}$, with  $\textbf{d}_t=\{d_t^{1},\ldots,d_t^{q}\}$, $t=1,\ldots,r$, has $q$ input attributes 
and one output attribute $o_t\in[0,1]$. We also assume that dataset $ D $ has been split into two datasets $ R $ and $ T $ for training and testing respectively. The dataset $ D_T $, which contains the last $0.2\cdot r$ samples of $D$ (20\% of $D$), will not be used in the learning process and will only be used when it ends at prediction time. 

The first step of EFS-LSTM-MOEA (line 1) is to split the training dataset $ R $ into $ n $ partitions $R_k=\big\{\{\textbf{d}_1^k,o_1^k\},\ldots,\{\textbf{d}_{l}^k,o_l^k\}\big\}$, $\textbf{d}_t^k=\{d_t^{k1},\ldots,d_t^{kq}\}$, $ k=1,\ldots,n $, $t=1,\ldots,l$, of equal size $l=r/n$ and with consecutive samples in time, such that each partition $ R_k $ contains samples at times earlier than the samples in the partition $ R_{k+1} $, for all $ k=1,\ldots,n-1 $.

In the next step (line 2), a MOEA is run to solve an multi-objetive optimization problem of $ n $ objective functions $\mathcal{F}_{R_k}:\mathds{B}^q\times \mathds{R}^z \rightarrow \mathds{R}$, one for each dataset $ R_k $, $ k=1,\ldots,n $, expressed as follows:

\begin{equation}
\label{eq:problem}
Minimize \ \ \mathcal{F}_{R_k}(\textbf{x}), \ k=1,\ldots,n
\end{equation}

\noindent where \textbf{x} is a set of decision variables composed of the following two subsets of decision variables:

\begin{itemize}
	\item[--] $\textbf{s} = \{s_1,\ldots,s_q\}$, $s_i\in \mathds{B}=\{\mbox{true},\mbox{false}\}$,  $i=1,\ldots,q$. Each decision variable $ s_i $ indicates whether the input attribute $i$  is selected ($\mbox{true}$) or not ($\mbox{false}$) in the learning process.
	\item[--] $\textbf{w}=\{w_1,\ldots,w_z\}$, $w_i\in \mathds{R}$, $i=1,\ldots,z$, with:
	
	\begin{equation}
	z = 4\cdot(q\cdot u + u^2 + 2\cdot u) + u + 1
	\end{equation}
	
	\noindent where $u$ is the number of units in the hidden layer\footnote{In this paper we consider, without loss of generality, an architecture with only one hidden layer.}.
\end{itemize}

The objective functions $\mathcal{F}_{R_k}(\textbf{x})$,  $k=1,\ldots,n$, return the \textit{root mean squared error} (RMSE) of an LSTM with the weights and biases of the set of decision variables \textbf{w}, selecting the attributes with value $\mbox{true}$ in the set of decision variables \textbf{s}, for the $ l $ samples of the dataset $ R_k $. 

\begin{figure*}[!t]
	\centering
	\resizebox{0.9\textwidth}{!}{\includegraphics[width=\textwidth]{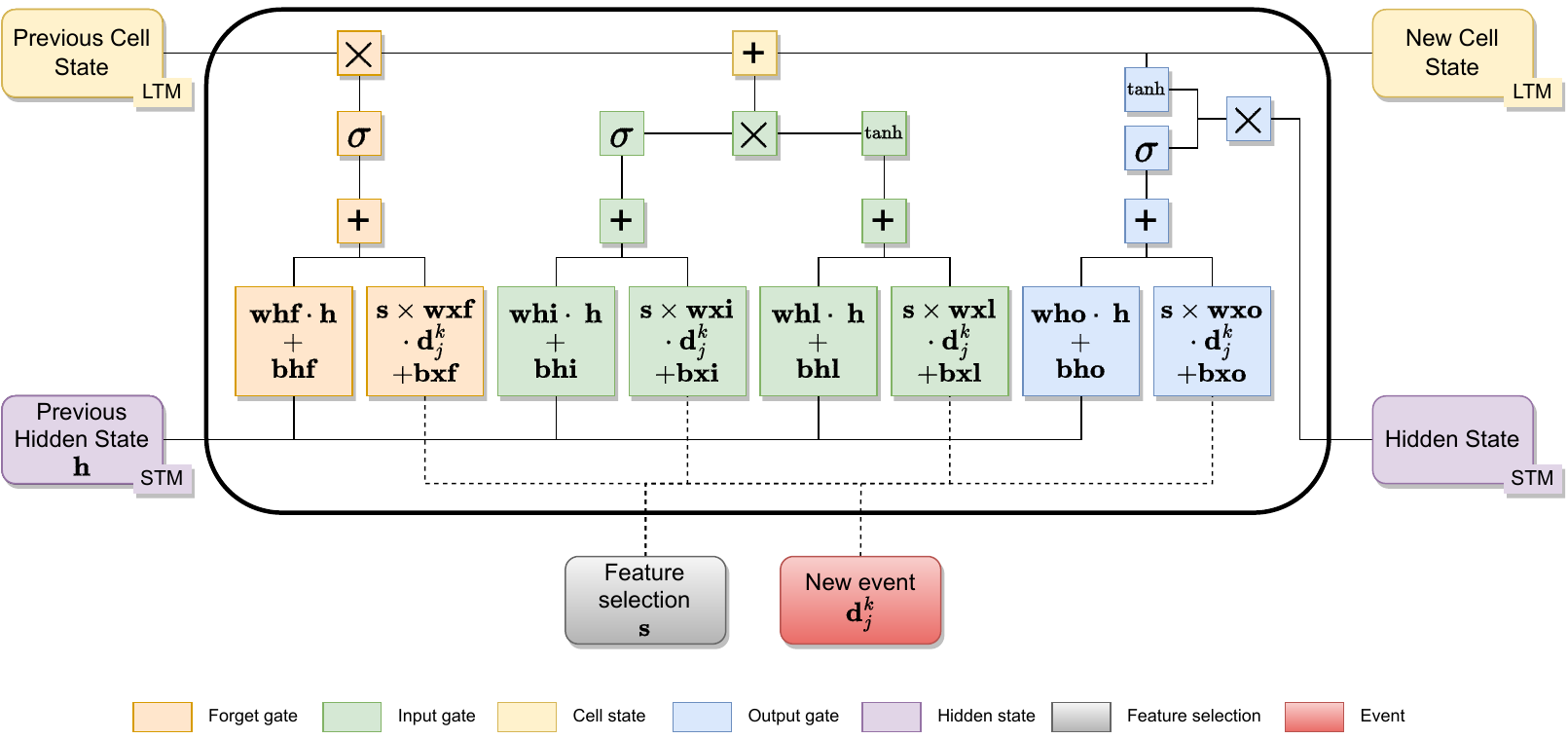}}
	\caption{Embedded feature selection in an LSTM network.}
	\label{fig:LSTM}
\end{figure*}

The calculation of the objective function $\mathcal{F}_{R_k}(\textbf{x})$,  $k=1,\ldots,n$ consists of the LSTM \textit{forward pass} with the weights and biases in \textbf{w} in which the feature selection established in \textbf{s} has been embedded, to later calculate the RMSE of the predictions obtained. Algorithm \ref{alg:ObjectiveFunction} shows the pseudo-code\footnote{In Algorithm \ref{alg:ObjectiveFunction}, the operator $\times$ is the \textit{element-wise product}.} required for the calculation of the objective function $\mathcal{F}_{R_k}$. Algorithm \ref{alg:ObjectiveFunction} requires using the function Convert($ \textbf{w} $) (Algorithm \ref{alg:Convert}) to convert the set \textbf{w} into the \textit{event} and \textit{hidden state} weight and bias matrices of the LSTM, with their corresponding dimensions, for the \textit{forget gate}, \textit{ignore input gate}, \textit{learn input gate} and \textit{output gate}, plus the weight matrix and the bias matrix of the final, fully connected layer, which takes the LSTM output and transforms it to our desired output size (a total of 9 weight matrices and 9 bias matrices, whose names and dimensions are shown in Algorithm \ref{alg:Convert}). In turn, Algorithm \ref{alg:Convert} uses the function Extract($\textbf{w},i,d_1,d_2 $) to extract $d_1\cdot d_2$ real numbers of set $ \textbf{w} $ from index $ i $ and construct a  $d_1 \times d_2$ matrix of weights or biases. The proposed LSTM architecture with embedded FS is shown graphically in Figure \ref{fig:LSTM}.

\begin{algorithm*}[!t]
	\caption{\textit{Function $ \mathcal{F}_{R_k} $}}
	\label{alg:ObjectiveFunction}
	\begin{algorithmic}[1]
		\REQUIRE $\textbf{x}=\{\textbf{s},\textbf{w}\}$ \COMMENT{Set of decision variables of the problem (\ref{eq:problem})}
		\REQUIRE $R_k=\big\{\{\textbf{d}_1^k,o_1^k\},\ldots,\{\textbf{d}_{l}^k,o_l^k\}\big\}$ \COMMENT{Training dataset}
		\ENSURE $F$ \COMMENT{Evaluation of the set of decision variables $ \textbf{x}$ on training dataset $R_k$}
		
		\STATE $\textbf{wxi}, \textbf{wxf}, \textbf{wxl}, \textbf{wxo}, \textbf{bxi}, \textbf{bxf}, \textbf{bxl}, \textbf{bxo}, 
		\textbf{whi}, \textbf{whf}, \textbf{whl}, \textbf{who}, \textbf{bhi}, \textbf{bhf}, \textbf{bhl}, \textbf{bho}, \textbf{wo}, \textbf{bo}
		\leftarrow$ Convert($\textbf{w}$)	

		\STATE $\textbf{h},\textbf{c}\leftarrow \textbf{0}$ \COMMENT{Initialize cell and hidden states with zeros}
		
		\FOR{$j=1$ \TO $l$} 
		\STATE $\textbf{f}\leftarrow \sigma\big((\textbf{whf}\cdot \textbf{h} + \textbf{bhf}) + (\textbf{s} \times\textbf{wxf}\cdot \textbf{d}_j^k + \textbf{bxf})\big) \times \textbf{c}$ \COMMENT{Forget gate}
		\STATE $\textbf{i}\leftarrow \sigma\big((\textbf{whi}\cdot \textbf{h} + \textbf{bhi}) + (\textbf{s} \times\textbf{wxi}\cdot \textbf{d}_j^k + \textbf{bxi})\big)
		\times
		\tanh\big((\textbf{whl}\cdot \textbf{h} + \textbf{bhl}) + (\textbf{s} \times\textbf{wxl}\cdot \textbf{d}_j^k + \textbf{bxl})\big)
		$ \COMMENT{Input gate}
		\STATE $\textbf{c}\leftarrow\textbf{f}+\textbf{i}$ \COMMENT{Cell state}
		\STATE $\textbf{h}\leftarrow
		\sigma\big((\textbf{who}\cdot \textbf{h} + \textbf{bho}) + (\textbf{s} \times\textbf{wxo}\cdot \textbf{d}_j^k + \textbf{bxo})\big)\times\tanh\big(\textbf{c}\big)$ \COMMENT{Output gate}
		\STATE ${y}_j^k\leftarrow\textbf{wo}\cdot \textbf{h}+\textbf{bo}$ \COMMENT{Model output for event $\textbf{d}_j^k$}
		\ENDFOR
		\STATE $F\leftarrow$ RMSE\big($({y}_1^k,\ldots,{y}_l^k),(o_1^k\ldots,o_l^k)$\big) \COMMENT{LSTM performance evaluation}
		\RETURN {$F$} 
	\end{algorithmic}
\end{algorithm*}

\begin{algorithm*}[!t]
	\caption{Convert}
	\label{alg:Convert}
	\begin{algorithmic}[1]
		\REQUIRE $\textbf{w}=\{w_1,\ldots,w_z\}, \ w_i\in \mathds{R}, \ i=1,\ldots,z$ \COMMENT{Subset $ \textbf{w} $ of the set of decision variables of the problem (\ref{eq:problem})}
		\ENSURE $\textbf{wxi}, \textbf{wxf}, \textbf{wxl}, \textbf{wxo}, \textbf{bxi}, \textbf{bxf}, \textbf{bxl}, \textbf{bxo}, 
		\textbf{whi}, \textbf{whf}, \textbf{whl}, \textbf{who}, \textbf{bhi}, \textbf{bhf}, \textbf{bhl}, \textbf{bho}, \textbf{wo}, \textbf{bo}$ \COMMENT{LSTM weight and bias matrices}
		
		\STATE $\textbf{wxi}\leftarrow$ Extract($\textbf{w},1,u,q$)	
		\COMMENT{Event weights for ignore input gate, $u\times q$ matrix}
		
		\STATE $\textbf{wxf}\leftarrow$ Extract($\textbf{w},q\cdot u+1,u,q$)	\COMMENT{Event weights for forget gate, $u\times q$ matrix}
		
		\STATE $\textbf{wxl}\leftarrow$ Extract($\textbf{w},2\cdot q\cdot u+1,u,q$) \COMMENT{Event weights for learn input gate, $u\times q$ matrix}
		
		\STATE $\textbf{wxo}\leftarrow$ Extract($\textbf{w},3\cdot q\cdot u+1,u,q$) \COMMENT{Event weights for output gate, $u\times q$ matrix}
		
		\STATE $\textbf{bxi}\leftarrow$ Extract($\textbf{w},4\cdot q\cdot u+1,1,u$) \COMMENT{Event biases for ignore input gate, $1\times u$ matrix}
		
		\STATE $\textbf{bxf}\leftarrow$ Extract($\textbf{w},4\cdot q\cdot u+u+1,1,u$) \COMMENT{Event biases for forget gate, $1\times u$ matrix}
		
		\STATE $\textbf{bxl}\leftarrow$ Extract($\textbf{w},4\cdot q\cdot u+2\cdot u+1,1,u$) \COMMENT{Event biases for learn input gate, $1\times u$ matrix}
		
		\STATE $\textbf{bxo}\leftarrow$ Extract($\textbf{w},4\cdot q\cdot u+3\cdot u+1,1,u$) \COMMENT{Event biases for output gate, $1\times u$ matrix}
		
		\STATE $\textbf{whi}\leftarrow$ Extract($\textbf{w},4\cdot q\cdot u+4\cdot u+1,u,u$) \COMMENT{Hidden state weights for ignore input gate, $u\times u$ matrix}
		
		\STATE $\textbf{whf}\leftarrow$ Extract($\textbf{w},4\cdot q\cdot u+4\cdot u+u^2+1,u,u$) \COMMENT{Hidden state weights for forget gate, $u\times u$ matrix}
		
		\STATE $\textbf{whl}\leftarrow$ Extract($\textbf{w},4\cdot q\cdot u+4\cdot u+2\cdot u^2+1,u,u$)	\COMMENT{Hidden state weights for learn input gate, $u\times u$ matrix}		
		
		\STATE $\textbf{who}\leftarrow$ Extract($\textbf{w},4\cdot q\cdot u+4\cdot u+3\cdot u^2+1,u,u$)	\COMMENT{Hidden state weights for output gate, $u\times u$ matrix}	
		
		\STATE $\textbf{bhi}\leftarrow$ Extract($\textbf{w},4\cdot q\cdot u+4\cdot u+4\cdot u^2+1,1,u$)	\COMMENT{Hidden state biases for ignore input gate, $1\times u$ matrix}	
		
		\STATE $\textbf{bhf}\leftarrow$ Extract($\textbf{w},4\cdot q\cdot u+4\cdot u+4\cdot u^2+u+1,1,u$) \COMMENT{Hidden state biases for forget gate, $1\times u$ matrix}	
		
		\STATE $\textbf{bhl}\leftarrow$ Extract($\textbf{w},4\cdot q\cdot u+4\cdot u+4\cdot u^2+2\cdot u+1,1,u$)	\COMMENT{Hidden state biases for learn input gate, $1\times u$ matrix}	
		
		\STATE $\textbf{bho}\leftarrow$ Extract($\textbf{w},4\cdot q\cdot u+4\cdot u+4\cdot u^2+3\cdot u+1,1,u$)	\COMMENT{Hidden state biases for output gate, $1\times u$ matrix}	
		
		\STATE $\textbf{wo}\leftarrow$ Extract($\textbf{w},4\cdot q\cdot u+4\cdot u+4\cdot u^2+4\cdot u+1,1,u$) \COMMENT{Final, fully connected layer weights, $1\times u$ matrix}
		
		\STATE $\textbf{bo}\leftarrow$ Extract($\textbf{w},4\cdot q\cdot u+4\cdot u+4\cdot u^2+4\cdot u+u+1,1,1$) \COMMENT{Final, fully connected layer biases, $1\times 1$ matrix}
		
		\RETURN {$\textbf{wxi}, \textbf{wxf}, \textbf{wxl}, \textbf{wxo}, \textbf{bxi}, \textbf{bxf}, \textbf{bxl}, \textbf{bxo}, 
			\textbf{whi}, \textbf{whf}, \textbf{whl}, \textbf{who}, \textbf{bhi}, \textbf{bhf}, \textbf{bhl}, \textbf{bho}, \textbf{wo}, \textbf{bo}$} 
	\end{algorithmic}
\end{algorithm*}

The solution of the multi-objective optimization problem of Eq. (\ref{eq:problem}) is a set of \textit{non-dominated solutions} $\big\{M_1,\ldots,M_m\big\}$ each of which corresponds to an LSTM network.  This set of LSTM networks is used by the EFS-LSTM-MOEA algorithm (lines 3 and 4) to build an ensemble machine learning model with the \textit{stacking} method. Finally, the ensemble prediction model is used to perform multi-step-ahead predictions which are evaluated on the test dataset $ T $ (lines 5 y 6). Below are the details of the MOEA used by EFS-LTSM-MOEA, the ensemble learning stacking method and the technique used for multi-step ahead forecasting.

\subsection{The multi-objective evolutionary algorithm}
\label{MOEA}
We have used the \textit{Platypus}  platform \cite{platypus} to implement the MOEA. Although we could have used any other algorithm implemented in \textit{Platypus}, we have used \textit{NSGA-II} \cite{Deb02} because of its popularity and its good balance between effectiveness and efficiency. \textit{NSGA-II} creates a new population by combining $P$ parents and $P$ offspring and selecting the best $P$ solutions using first \textit{Pareto front dominance} and then \textit{crowding distance}, where $P$ is the population size. \textit{NSGA-II} uses a \textit{binary tournament selection} in which individuals are also compared using Pareto front dominance and crowding distance.

\begin{figure*}[!t]
	\centering
	\resizebox{0.9\textwidth}{!}{\includegraphics[width=\textwidth]{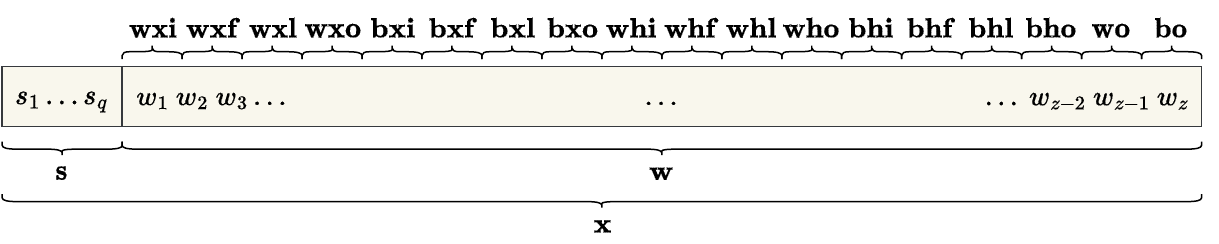}}
	\caption{Representation of an individual $\textbf{x}$.}
	\label{fig:representation}
\end{figure*}

To solve problem (\ref{eq:problem}), we have used a mixed binary-real encoding to represent the individuals in the population. Thus, an individual \textbf{x} is represented by two components $ \{\textbf{s},\textbf{w}\} $ as in problem (\ref{eq:problem}). The \textbf{s} component is represented as a \textit{bit string}, while the \textbf{w} component is represented as an array of \textit{floating point} numbers. Figure \ref{fig:representation} graphically illustrates the representation of individuals used by the MOEA.

Similarly, the fitness function of an individual $ \textbf{x} $ for each objective function is the function $\mathcal{F}_{R_k}(\textbf{x})$, $k=1,\ldots,n$, of the problem (\ref{eq:problem}) calculated with Algorithm \ref{alg:ObjectiveFunction}. We use \textit{half uniform crossover} \cite{ESHELMAN1991}  and \textit{bit flip mutation} \cite{davis1991} for the  binary-valued components of $ \textbf{s} $, and
\textit{simulated binary crossover} \cite{Deb1995SimulatedBC} and \textit{polynomial mutation} \cite{Deb1996ACG} for the  real-valued components of $ \textbf{w} $. These operators are set by default in the \textit{Platypus} platform for binary and real representations respectively.

\subsection{Stacking-based ensemble learning}
\label{Stacking}

\textit{Ensemble learning} refers to a machine learning technique where multiple models, known as base learners or weak learners, are combined to form a stronger and more accurate model. The idea behind ensemble learning is that by combining the predictions of multiple models, the resulting ensemble model can make more robust and accurate predictions than any individual model.
\textit{Stacking-based ensemble learning}, also known as \textit{stacked generalization}, is a specific type of ensemble learning method. In stacking, multiple diverse base learners are trained on the same dataset, and their predictions are then combined using another model called a \textit{meta-learner}. The meta-learner takes the predictions of the base learners as input and learns to make the final prediction.

The stacking process typically involves two phases. In the first phase, the base learners are trained independently on the training data, generating predictions. In the second phase, the meta-learner is trained on the first phase's predictions to make the final prediction. The idea behind stacking is to leverage the diverse strengths and weaknesses of the base learners, allowing the meta-learner to learn a combination of their predictions that maximizes the overall performance. Stacking-based ensemble learning can be more powerful than individual models or traditional ensemble methods because it learns to effectively weigh and combine the predictions of the base learners, taking advantage of their complementary strengths. It has been successfully applied in various machine learning tasks and is particularly useful when dealing with complex problems where no single model can provide satisfactory results.

Stacking-based ensemble learning can increase \textit{generalization} in machine learning models. Generalization refers to the ability of a model to perform well on unseen data or data from the real world. Stacking achieves improved generalization by combining the predictions of multiple base learners, which helps to reduce the individual errors or biases of each model.
The diversity of the base learners is a key factor in improving generalization.  This diversity allows the ensemble model to cover a wider range of potential solutions and reduces the risk of overfitting to specific patterns in the training data. 

In our proposal, stacking-based ensemble learning is treated differently from the usual approach. While in the usual approach different base learners are trained on the same dataset during the first phase, in our approach the same base learner (LSTM) is trained on different splits of the same time series dataset distributed in time using a multi-objective optimization approach as in problem (\ref{eq:problem}), thus obtaining multiple LSTM models which are assembled in the second phase of the stacking process. This process allows, like the usual stacking approach, to capture different aspects of the data and learn different patterns, improving the generalization. Below we describe the proposed process to build the stacking-based ensemble learning model from the set $\big\{M_1,\ldots,M_m\big\}$ of LSTM models found by the MOEA.

\begin{enumerate}
	
	\item Evaluate each input data $\textbf{d}_t$, $ t = 1, \ldots, r $, in each LSTM model $ M_j $, $ j = 1, \ldots, m $. The model outputs $Y_{M_j}(\textbf{d}_t)$, $ t = 1, \ldots, r $, $ j = 1, \ldots, m $, are obtained in this step.

	\item Build a dataset $ E_R $  with the model outputs  $Y_{M_j}(\textbf{d}_t)$, $ t = 1, \ldots, r $, $ j = 1, \ldots, m $ and the corresponding observations $o_t$, $ t = 1, \ldots, r $,  as follows:
	\begin{equation*}
			E_R=\begin{pmatrix}
					Y_{M_1}(\textbf{d}_1)& Y_{M_2}(\textbf{d}_1) & \ldots & Y_{M_m}(\textbf{d}_1) & o_1 \\
					Y_{M_1}(\textbf{d}_2)& Y_{M_2}(\textbf{d}_2) & \ldots & Y_{M_m}(\textbf{d}_2) & o_2 \\
					\vdots & \vdots & \ddots & \vdots & \vdots \\
					Y_{M_1}(\textbf{d}_r)& Y_{M_2}(\textbf{d}_r) & \ldots & Y_{M_m}(\textbf{d}_r) & o_{r} \\		
			\end{pmatrix}
			\end{equation*}	
	\item Finally, a prediction model $\Psi$ is built using the dataset $E_R$ as training data and some learning algorithm $\Phi$ for regression.
	
\end{enumerate}

\subsection{Multi-step-ahead forecasting}
\label{Multistep}
\textit{Multi-step ahead time series forecasting} \cite{BENTAIEB20127067}, also known as \textit{multi-step forecasting} or \textit{horizon forecasting}, is the process of predicting multiple future data points in a time series beyond the next time step. Instead of just forecasting the next single data point, the objective is to make predictions for a specified number of steps into the future. With multi-step ahead time series forecasting, the next $h$ values $y_{r+1},\ldots , y_{r+h}$ are forecast from a historical time series $y_{1},\ldots, y_{r}$ composed of $r$ samples, where $h > 1$ is the forecasting horizon.

In this work we use the \textit{recursive strategy} \cite{7064712} for multi-step ahead time series forecasting. The recursive strategy for multi-step time series forecasting involves making step-by-step predictions into the future. In this approach, one step ahead is predicted at a time, and then the predicted value is used as input for the next prediction. This process is repeated iteratively for the desired number of steps ahead.
The performance of the model can be evaluated with the recursive strategy using the training set $ R $ or the test set $ T $. Although we could have used other strategies for multi-step ahead time series forecasting, in this work we have preferred to use the recursive strategy for its simplicity and easy implementation in all the learning algorithms compared in this paper, and as a first approximation for its inclusion in the EFS-LSTM-MOEA algorithm.

\subsection{Feature importance}
\label{FI}

The importance of the attributes can be obtained by calculating the frequency of appearance of each attribute in the set of non-dominated forecast models $ \{M_1,\ldots,M_m\} $ found with the MOEA, using the following formula:

\begin{equation}
I_i=\frac{1}{m}\cdot\displaystyle\sum_{j=1}^m s_i^j, \ i=1,\ldots,q
\end{equation}

\noindent where $s_i^j\in\{0,1\}$ indicates whether attribute $ i $ has been selected or not in model $ M_j $, $i=1,\ldots,q$, $j=1,\ldots,m$. In this way, an attribute $i$ that has been selected in all the $m$ non-dominated prediction models identified by the MOEA will obtain an importance $I_i=1$, while if it has not been selected by any prediction model it will obtain an importance $I_i=0$.

\section{Experiments and results} 
\label{ER}

This section describes the set of experiments that have been carried out to verify the performance of the proposed method, the datasets and algorithms used for comparisons, and shows the results obtained.

\subsection{Datasets}
Two air quality datasets have been used. The first dataset (Table \ref{tab:sum-aq}) has been obtained from the \textit{UCI Machine Learning Repository} \cite{Dua2019} and contains hourly air quality data from an Italian city. It has 12 attributes and 1000 instances that correspond to measurements made between February 21, 2005 and April 4, 2005. The attribute to predict is $NO_X$. The second dataset (Table \ref{tab:sum-lorca}) corresponds to daily air quality measurements from a monitoring station located in Lorca, a city in the Region of Murcia (Spain). The data have been provided by Consejería de Medio Ambiente, Mar Menor, Universidades e Investigación\footnote{https://sinqlair.carm.es/calidadaire/}.  There are a total of 1000 instances since the measurements were carried out between April 7, 2019 and December 31, 2021. The dataset has 10 attributes with the target being $NO_2$.

Both datasets have been preprocessed as follows: in order to eliminate missing values, a \textit{linear interpolation} has been performed; after this, a \textit{sliding window} process \cite{brownlee2020} with a window size of 3 has been applied to the datasets; afterwards, the datasets have been divided into two partitions, 80\% for training and the remaining 20\% for testing (as described in section \ref{GM}); finally, the partitions have been normalized separately to prevent data leakage.

\begin{table}
	\centering
	\resizebox{0.45\textwidth}{!}{
		\begin{tabular}{lccccc}
			\hline
			\textbf{Name} & \textbf{Units} &    \textbf{Min} &     \textbf{Mean} &     \textbf{Max} &     \textbf{Std} \\
			\hline
			$ CO(GT) $        & $mg/m^3$ & 	0.10 &    1.95 &     7.50 &   1.32 \\
			$ PT08.S1(CO) $   & $mg/m^3$ &  715.00 &  1109.05 &  1818.00 &  196.73 \\
			$ C6H6(GT) $      & $mg/m^3$ &  0.20 &     7.94 &    35.50 &    6.35 \\
			$ PT08.S2(NMHC) $ & $mg/m^3$ &  387.00 &   855.66 &  1675.00 &  247.69 \\
			$ PT08.S3(NOx) $  & $mg/m^3$ & 330.00 &   749.39 &  1804.00 &  227.35 \\
			$ NO2(GT) $       & $mg/m^3$ &   25.00 &   136.64 &   295.00 &   48.59 \\
			$ PT08.S4(NO2) $  & $mg/m^3$ &  551.00 &  1175.37 &  2147.00 &  273.88 \\
			$ PT08.S5(O3) $   & $mg/m^3$ &  221.00 &  1008.16 &  2159.00 &  413.70 \\
			$ T $             & ºC 		 &   -1.90 &    12.21 &    30.00 &    6.65 \\
			$ RH $            & \% 	     &    9.90 &    51.73 &    86.60 &   17.78 \\
			$ AH $            & \% 	     &    0.18 &     0.73 &     1.39 &    0.27 \\
			$ NOxGT $         & ppb &   25.00 &   288.60 &   959.00 &  180.40 \\		\hline& 
		\end{tabular}
	}
	\caption{Initial attribute statistics of the Italian city air quality dataset.}
	\label{tab:sum-aq}
\end{table}

\begin{table}
	\centering
	\resizebox{0.36\textwidth}{!}{
		\begin{tabular}{lccccc}
			\hline
			\textbf{Name} & \textbf{Units} & \textbf{Min} & \textbf{Mean} & \textbf{Max} & \textbf{Std} \\ \hline
			$NO$            & $\mu g/m^3$      & 1.00         & 2.56          & 8.00         & 0.98         \\
			$O_3$           & $\mu g/m^3$      & 5.00         & 44.60         & 120.00       & 16.11        \\
			$TMP$           & ºC               & 5.00         & 19.23         & 34.00        & 6.26         \\
			$HR$            & \% R.H.          & 30.00        & 71.64         & 100.00       & 16.08        \\
			$NO_X$          & $\mu g/m^3$      & 5.00         & 13.16         & 32.00        & 4.65         \\
			$DD$            & degrees          & 37.00        & 208.66        & 309.00       & 47.25        \\
			$RS$            & $W/m^3$          & 2.00         & 158.79        & 345.00       & 72.13        \\
			$VV$            & $m/s$            & 0.00         & 0.66          & 5.00         & 0.53         \\
			$PM_{10}$       & $\mu g/m^3$      & 3.00         & 23.74         & 172.00       & 15.11        \\
			$NO_2$          & $\mu g/m^3$      & 3.00         & 9.23          & 23.00        & 3.58         \\ \hline
		\end{tabular}
	}
	\caption{Initial attribute statistics of the Lorca air quality dataset.}
	\label{tab:sum-lorca}
\end{table}

\subsection{Description of the experiments and results}
\begin{figure}[!t]
	\centering
	\vskip -1cm
	\resizebox{0.45\textwidth}{!}{\includegraphics[width=\textwidth]{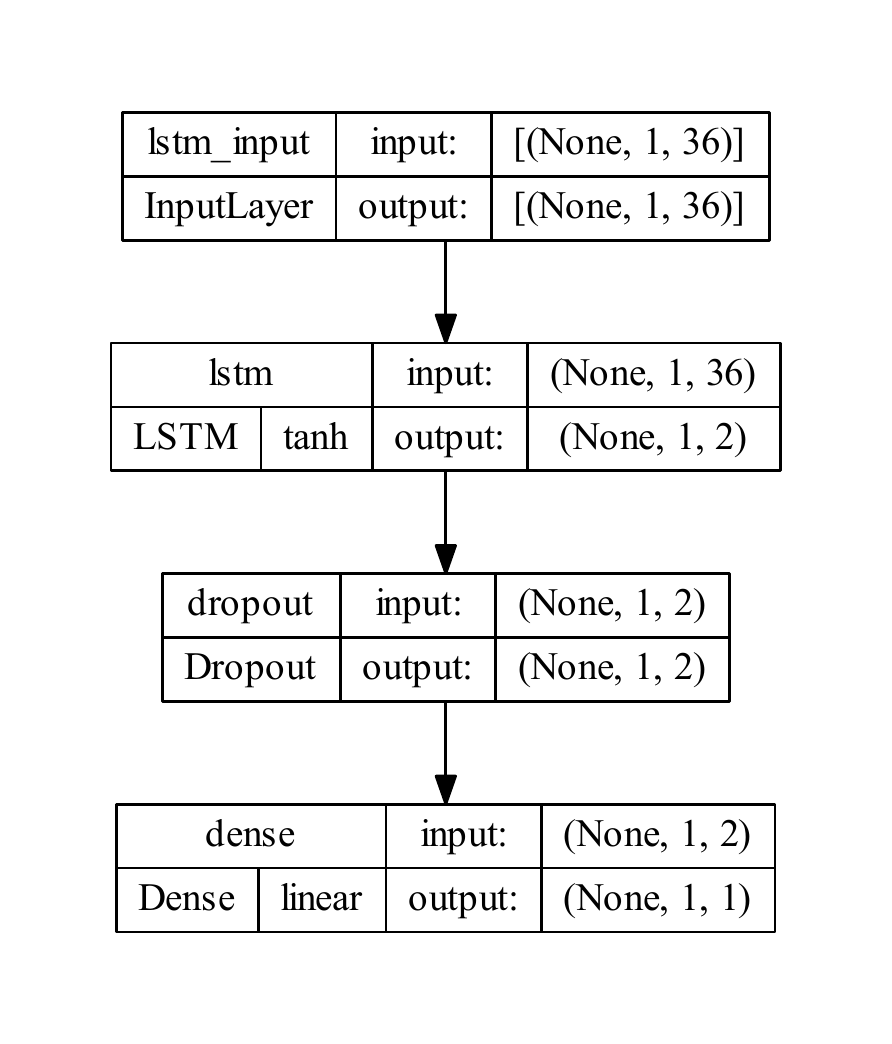}}
	\vskip -1cm
	\caption{Architecture of the conventional LSTM network for the Italy city air quality problem.}
	\label{fig:architectureLSTM-Italy}
\end{figure}

\begin{figure}[!ht]
	\centering
	\vskip -1cm
	\resizebox{0.45\textwidth}{!}{\includegraphics[width=\textwidth]{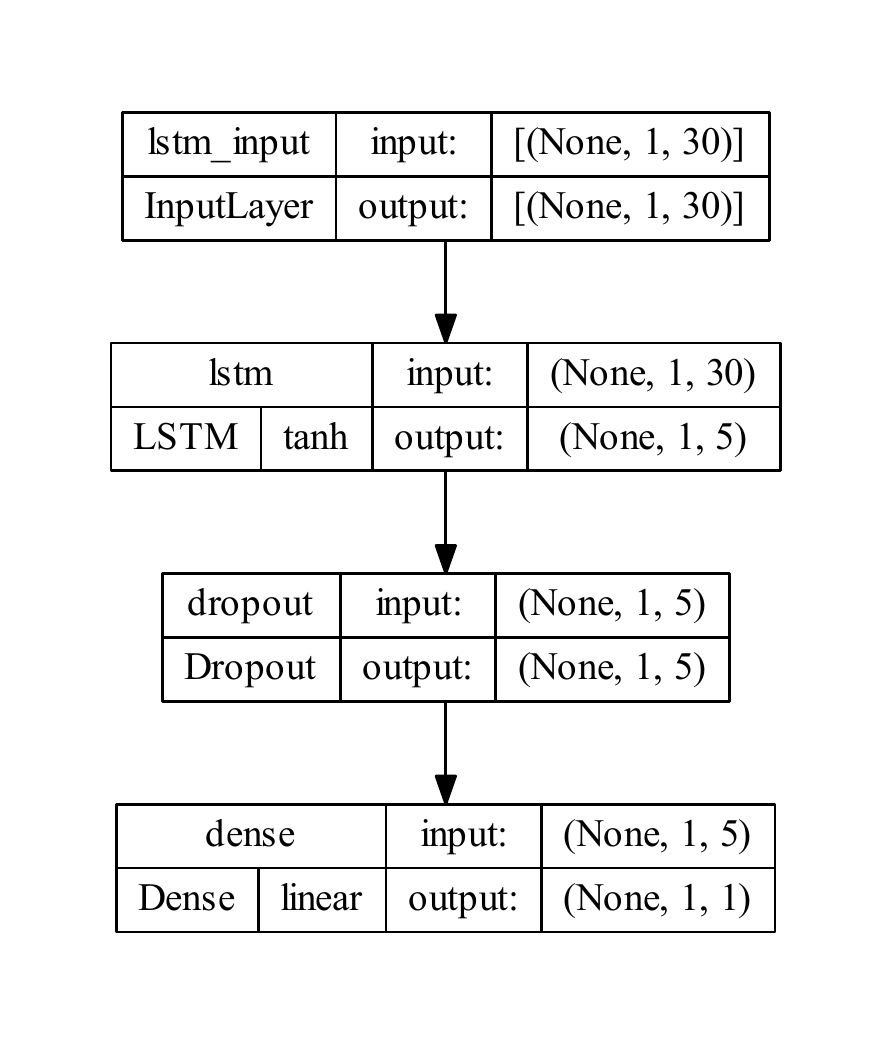}}
	\vskip -1cm
	\caption{Architecture of the conventional LSTM network for the Lorca air quality problem.}
	\label{fig:architectureLSTM-Lorca}
\end{figure}

The main goal of the experiments is to verify the generalization capacity of the proposed method, in comparison with a conventional LSTM network. To do this, in a first block of experiments the following steps have been carried out:

\begin{enumerate}
	\item The optimization problem (\ref{eq:problem}) has been set with $n=5$ \textit{objective functions}, that is, the training data has been divided into 5 partitions\footnote{The number of partitions $n=5$ has been chosen after previous experimentation with $n\in\{2,3,4,5\}$.}. EFS-LSTM-MOEA has been run \textit{10 times} with different seeds, with a \textit{population size} of $ 50 $ individuals, $ 50,000 $ \textit{generations}  and  \textit{crossover and mutation probabilities} equal to $ 1.0 $. The \textit{learning algorithm} $\Phi$ used to build the ensemble model with the non-dominated prediction models found with the MOEA has been \textit{random forest} \cite{breiman2001random}.
	\item The conventional LSTM network has been configured with a hidden layer followed by a \textit{dropout} layer. The hyper-parameters have been tuned using \textit{grid search}, with $u\in\{2, 5, 10\}$, $epochs \in\{ 100, 500, 1000\}$, $bach\_size \in\{8, 16, 32, 128\}$, where models are evaluated in \textit{$3$-fold cross-validation}. The optimized hyper-parameters were $ u= 2 $, $ epoch=1000 $ and $ batch\_size=32 $ for the Italy city dataset, and $ u= 5 $, $ epoch=500 $ and $ batch\_size=128 $ for the Lorca dataset. Figures \ref{fig:architectureLSTM-Italy}  and  \ref{fig:architectureLSTM-Lorca}  show the architecture of the conventional LSTM network after hyper-parameter tuning for the Italy city and Lorca air quality problems respectively. Then, with the selected hyper-parameters, 10 conventional LSTM networks have been built using different seeds for the \textit{Glorot uniform initializer} of weights and biases.
\end{enumerate}

In a second block of experiments,  the EFS-LSTM-MOEA method has been compared with other state-of-the-art FS methods that can be embedded in LSTM networks. We have experimented with two of them: CancelOut \cite{cancelout}  and  EAR-FS \cite{XUE2023111084}.  The details of both methods and their configurations are described below:

\begin{enumerate}
	\item CancelOut is an artificial neural network layer located between the input layer and the  hidden layer. Each neuron in the CancelOut layer is connected to a particular input and fully connected to the neurons in the hidden layer. In this way, the values after the activation function in the CancelOut layer indicate contributions to the output of a corresponding input attribute. We have added a CancelOut layer in the architecture used above for the conventional LSTM network. 
	The method has been run $10$ times with different seeds\footnote{The CancelOut code is available at \url{https://github.com/unnir/CancelOut}.}.
	
	\item EAR-FS is composed of three modules. The first module is an MLP with embedded attention to learn the feature weights. The second module processes the feature weights and obtains a feature ranking. The third module generates appropriate subsets of attributes based on feature ranking and some classifier. EAR-FS was implemented by its authors for classification tasks and therefore had to be adapted in this paper for regression tasks. The hyper-parameters of EAR-FS were established according to the recommendations of the authors in \cite{XUE2023111084}, i.e. $ momentum = 0.9 $, $ weight \ decay = 0.00001 $, $minimum \ value \ of \ the \ learning \ rate = 0.001$, $ M = 50 $, $ L = 25 $ and $ K_p= 0.1$. In the third module of EAR-FS we have used the LSTM networks of the Figures \ref{fig:architectureLSTM-Italy}  and  \ref{fig:architectureLSTM-Lorca} for subset generation. Like the rest of the methods compared in this paper, EAR-FS has been run 10 times with different seeds\footnote{The EAR-FS code is available at \url{https://github.com/xueyunuist/EAR-FS}.}.

\end{enumerate}
%
%
%

%
%
%
%
%
%
%
%
%

Tables \ref{tab:aq-FS-MOEA} to \ref{tab:lorca-EAR-FS} show the results obtained with EFS-LSTM-MOEA, conventional LSTM, CancelOut and EAR-FS for the Italy city and Lorca air quality problems. For each of the 10 runs, the average RMSE of the forecast models is shown over a 3-steps ahead horizon. The average, maximum and minimum results of the 10 runs are also shown.


\begin{table}[!h]
	\centering
	\resizebox{0.485\textwidth}{!}{
		\begin{tabular}{cccc}
			\hline
			\textbf{Run}  & \textbf{\begin{tabular}[c]{c} Average RMSE  \\ in training\end{tabular}} & \textbf{\begin{tabular}[c]{c}Average RMSE \\ in test\end{tabular}} & \textbf{\begin{tabular}[c]{c} Average  number \\ of attributes\end{tabular}}\\ \hline
			1               & 0.11361             & 0.11478            & 18.65 \\
			2               & 0.10792             & 0.11319            & 22.46 \\
			3               & 0.11220             & 0.11774            & 21.47  \\
			4               & 0.14361             & 0.11682            & 17.42  \\
			5               & 0.11505             & 0.11299            & 25.11  \\
			6               & 0.11036             & 0.09682            & 17.68  \\
			7               & 0.14128             & 0.11237            & 20.95  \\
			8               & 0.16157             & 0.11282            & 19.45  \\
			9               & 0.13052             & 0.11472            & 27.36  \\
			10               & 0.10932             & 0.09823            & 17.84  \\ \hline
			\textbf{Average}   & 0.12454             & 0.11105            & 20.84  \\
			\textbf{Min}    & 0.10792             & 0.09682            & 17.42  \\
			\textbf{Max}    & 0.16157             & 0.11774            & 27.36  \\ \hline
		\end{tabular}
	}	
	\caption{Evaluation of EFS-LSTM-MOEA for the Italy city air quality problem.}
	\label{tab:aq-FS-MOEA}
\end{table}

\begin{table}[!h]
	\centering
	\resizebox{0.485\textwidth}{!}{
		\begin{tabular}{cccc}
			\hline
			\textbf{Run}  & \textbf{\begin{tabular}[c]{c} Average RMSE  \\ in training\end{tabular}} & \textbf{\begin{tabular}[c]{c}Average RMSE \\ in test\end{tabular}} & \textbf{\begin{tabular}[c]{c} Average  number \\ of attributes\end{tabular}}\\ \hline
			1               & 0.19022             & 0.18785            & 17.07           \\
			2               & 0.18666             & 0.17708            & 20.08           \\
			3               & 0.17906             & 0.17585            & 18.28           \\
			4               & 0.18340             & 0.18130            & 19.72           \\
			5               & 0.16558             & 0.17656            & 19.79           \\
			6               & 0.18373             & 0.18196            & 18.05           \\
			7               & 0.18236             & 0.17790            & 15.73           \\
			8               & 0.18386             & 0.18357            & 17.34           \\
			9               & 0.18772             & 0.17261            & 21.99           \\
			10               & 0.17950             & 0.18492            & 17.07           \\ \hline
			\textbf{Average}   & 0.18221             & 0.17996            & 18.51           \\
			\textbf{Min}    & 0.16558             & 0.17261            & 15.73           \\ 
			\textbf{Max}    & 0.19022             & 0.18785            & 21.99           \\ \hline
		\end{tabular}
	}
	\caption{Evaluation of EFS-LSTM-MOEA for the Lorca air quality problem.}
	\label{tab:lorca-FS-MOEA}
\end{table}

\begin{table}[!h]
	\centering
	\resizebox{0.345\textwidth}{!}{
	\begin{tabular}{ccc}
		\hline
		\textbf{Run}  & \textbf{\begin{tabular}[c]{c} Average RMSE  \\ in training\end{tabular}} & \textbf{\begin{tabular}[c]{c}Average RMSE \\ in test\end{tabular}} \\ \hline
		1             & 0.10024             & 0.22459            \\
		2             & 0.10419             & 0.14902            \\
		3             & 0.10065             & 0.17335            \\
		4             & 0.10047             & 0.20673            \\
		5             & 0.09668             & 0.16784            \\
		6             & 0.10079             & 0.15971            \\
		7             & 0.10039             & 0.20077            \\
		8             & 0.09829             & 0.16911            \\
		9             & 0.10118             & 0.19565            \\
		10             & 0.09642             & 0.17903            \\ \hline
		\textbf{Average} & 0.09993             & 0.18258            \\
		\textbf{Min}  & 0.09642             & 0.14902            \\ 
		\textbf{Max}  & 0.10419			    & 0.22459			 \\ \hline
	\end{tabular}}
	\caption{Evaluation of conventional LSTM for the Italy city air quality problem.}
	\label{tab:aq-conventional}
\end{table}

\begin{table}[!h]
	\centering
	\resizebox{0.345\textwidth}{!}{
		\begin{tabular}{ccc}
			\hline
			\textbf{Run}  & \textbf{\begin{tabular}[c]{c} Average RMSE  \\ in training\end{tabular}} & \textbf{\begin{tabular}[c]{c}Average RMSE \\ in test\end{tabular}} \\ \hline
		1               & 0.15176             & 0.21608            \\
		2               & 0.15087             & 0.21415            \\
		3               & 0.15469             & 0.21094            \\
		4               & 0.14874             & 0.21556            \\
		5               & 0.15561             & 0.21891            \\
		6               & 0.15388             & 0.20233            \\
		7               & 0.15228             & 0.22158            \\
		8               & 0.15300             & 0.22005            \\
		9               & 0.14862             & 0.21640            \\
		10               & 0.15039             & 0.22347            \\ \hline
		\textbf{Average}   & 0.15198             & 0.21595            \\
		\textbf{Min}    & 0.14862             & 0.20233            \\ 
		\textbf{Max}    & 0.15561			  &	0.22347  			\\ \hline
	\end{tabular}}
	\caption{Evaluation of conventional LSTM for the Lorca air quality problem.}
	\label{tab:lorca-conventional}
\end{table}

\begin{table}[!h]
	\centering
	\resizebox{0.485\textwidth}{!}{
		\begin{tabular}{cccc}
			\hline
			\textbf{Run}  & \textbf{\begin{tabular}[c]{c} Average RMSE  \\ in training\end{tabular}} & \textbf{\begin{tabular}[c]{c}Average RMSE \\ in test\end{tabular}} & \textbf{\begin{tabular}[c]{c} Average  number \\ of attributes\end{tabular}}\\ \hline
			1	& 0.08697	& 0.13645	& 16 \\
			2	& 0.08694	& 0.10763	& 18 \\
			3	& 0.08726	& 0.12010	& 16 \\
			4	& 0.08762	& 0.13535	& 20 \\
			5	& 0.08790	& 0.11363	& 17 \\
			6	& 0.08753	& 0.11458	& 18 \\
			7	& 0.08762	& 0.11783	& 13 \\
			8	& 0.08725	& 0.11431	& 18 \\
			9	& 0.08763	& 0.14273	& 20 \\
			10	& 0.08770	& 0.11312	& 16 \\\hline
			\textbf{Average}	& 0.08744	& 0.12157	& 17.20 \\
			\textbf{Min}	& 0.08694	& 0.10763	& 13 \\
			\textbf{Max}	& 0.08790	& 0.14273	& 20 
			\\ \hline
		\end{tabular}
	}
	\caption{Evaluation of CancelOut for the Italy city air quality problem.}
	\label{tab:aq-CancelOut}
\end{table}

\begin{table}[!h]
	\centering
	\resizebox{0.485\textwidth}{!}{
		\begin{tabular}{cccc}
			\hline
			\textbf{Run}  & \textbf{\begin{tabular}[c]{c} Average RMSE  \\ in training\end{tabular}} & \textbf{\begin{tabular}[c]{c}Average RMSE \\ in test\end{tabular}} & \textbf{\begin{tabular}[c]{c} Average  number \\ of attributes\end{tabular}}\\ \hline
			1	& 0.14671	& 0.19676	& 11 \\
			2	& 0.14721	& 0.19676	& 18 \\
			3	& 0.14684	& 0.19211	& 13 \\
			4	& 0.14653	& 0.19307	& 14 \\
			5	& 0.14606	& 0.19319	& 16 \\
			6	& 0.14660	& 0.19020	& 11 \\
			7	& 0.14697	& 0.19739	& 16 \\
			8	& 0.14735	& 0.20232	& 17 \\
			9	& 0.14648	& 0.20011	& 15 \\
			10	& 0.14697	& 0.19666	& 15 \\\hline
			\textbf{Average}	& 0.14677	& 0.19586	& 14.60 \\
			\textbf{Min}	& 0.14606	& 0.19020	& 11.00 \\
			\textbf{Max}	& 0.14735	& 0.20232	& 18.00 
			\\ \hline
		\end{tabular}
	}
	\caption{Evaluation of CancelOut for the Lorca air quality problem.}
	\label{tab:lorca-CancelOut}
\end{table}

\begin{table}[!h]
	\centering
	\resizebox{0.485\textwidth}{!}{
		\begin{tabular}{cccc}
			\hline
			\textbf{Run}  & \textbf{\begin{tabular}[c]{c} Average RMSE  \\ in training\end{tabular}} & \textbf{\begin{tabular}[c]{c}Average RMSE \\ in test\end{tabular}} & \textbf{\begin{tabular}[c]{c} Average  number \\ of attributes\end{tabular}}\\ \hline
1	& 0.17789	& 0.17539	& 33 \\
2	& 0.25570	& 0.26094	& 7 \\
3	& 0.22655	& 0.22957	& 32 \\
4	& 0.19613	& 0.19774	& 3 \\
5	& 0.16456	& 0.16616	& 28 \\
6	& 0.40532	& 0.40842	& 3 \\
7	& 0.21745	& 0.22096	& 5 \\
8	& 0.18941	& 0.19124	& 34 \\
9	& 0.30683	& 0.31251	& 1 \\
10	& 0.19719	& 0.20079	& 21 \\ \hline
\textbf{Average}	& 0.23370	& 0.23637	& 16.70 \\
\textbf{Min}	& 0.16456	& 0.16616	& 1.00 \\
\textbf{Max}	& 0.40532	& 0.40842	& 34.00\\ \hline
		\end{tabular}
	}
	\caption{Evaluation of EAR-FS for the Italy city air quality problem.}
	\label{tab:aq-EAR-FS}
\end{table}

\begin{table}[!h]
	\centering
	\resizebox{0.485\textwidth}{!}{
		\begin{tabular}{cccc}
			\hline
			\textbf{Run}  & \textbf{\begin{tabular}[c]{c} Average RMSE  \\ in training\end{tabular}} & \textbf{\begin{tabular}[c]{c}Average RMSE \\ in test\end{tabular}} & \textbf{\begin{tabular}[c]{c} Average  number \\ of attributes\end{tabular}}\\ \hline
1	& 0.17441	& 0.18871	& 18 \\
2	& 0.28826	& 0.28897	& 7 \\
3	& 0.27819	& 0.27920	& 4 \\
4	& 0.18378	& 0.18675	& 3 \\
5	& 0.17564	& 0.18021	& 23 \\
6	& 0.38046	& 0.38192	& 3 \\
7	& 0.32876	& 0.32850	& 5 \\
8	& 0.24851	& 0.23467	& 29 \\
9	& 0.32602	& 0.32614	& 1 \\
10	& 0.18299	& 0.19521	& 21 \\\hline
\textbf{Average}	& 0.25670	& 0.25903	& 11.40 \\
\textbf{Min}	& 0.17441	& 0.18021	& 1 \\
\textbf{Max}	& 0.38046	& 0.38192	& 29 
			\\ \hline
		\end{tabular}
	}
	\caption{Evaluation of EAR-FS for the Lorca air quality problem.}
	\label{tab:lorca-EAR-FS}
\end{table}

\section{Analysis of results and discussion}
\label{AnaResDis}
In this section we analyze and discuss the results obtained with EFS-LSTM-MOEA in comparison with the other three methods considered in this paper. The following highlights can be extracted from the results:

\begin{itemize}
\item[--] EFS-LSTM-MOEA has obtained better average, minimum and maximum RMSE on the test dataset than the conventional LSTM network for the two analyzed air quality problems, thus satisfying the main goal of this work. EFS-LSTM-MOEA has also outperformed the FS methods CancelOut and EAR-FS on the test dataset in the three indicators average, minimum and maximum, demonstrating greater generalization ability.

\item[--]  To analyze the generalization ability of the forecast models, we have calculated the \textit{overfitting ratio} of each model by dividing its RMSE on the training dataset by its RMSE on the test dataset. A ratio much less than 1 indicates overfitting. Table \ref{tab:overfitting} shows the average overfitting ratio of the 10 runs of EFS-LSTM-MOEA, conventional LSTM, CancelOut and EAR-FS for the two air quality problems analyzed and their average. The first notable result is that EFS-LSTM-MOEA has considerably improved the generalization ability of the conventional LSTM network which clearly shows overfitting. In fact, the results of EFS-LSTM-MOEA  on the test dataset are  almost the same as the results on the training dataset, with an overfitting ratio very close to 1. CancelOut also shows overfitting, although less than the conventional LSTM network, while EAR-FS practically obtains the same results on the test dataset and the training dataset, with an overfitting ratio slightly less than 1.

\begin{table}[!h]
	\centering
	\footnotesize
		\begin{tabular}{cccc}
			\hline
			\textbf{Method}  & \textbf{Italy city} & \textbf{Lorca} & \textbf{Average} \\ \hline
			EFS-LSTM-MOEA & {1.03295} & {0.99168} & {1.01236}\\
			Conventional LSTM & 0.49245 & 0.68765 & 0.59005 \\
			CancelOut & 0.70633 & 0.74648 & 0.72641\\
			EAR-FS & 0.99817 & 0.93080 & 0.96449\\ \hline
		\end{tabular}
	\caption{Average overfitting ratio of the 10 runs of the methods.}
	\label{tab:overfitting}
\end{table}

\item[--]  Statistical tests play a crucial role in the rigorous evaluation and comparison of machine learning algorithms. They provide a solid foundation for drawing reliable conclusions from experimental results, ensuring that findings are not merely due to chance or confounding factors. In this paper, the \textit{Diebold-Mariano} statistical test is used to compare the predictive accuracy of the forecasting models. It is particularly suitable for time series data, where observations are collected sequentially over time. The test helps determine whether one forecasting model significantly outperforms another in terms of forecasting accuracy for each of the $ h $ steps ahead. The Diebold-Mariano statistical test was performed to compare each of the four methods analyzed in this paper with the rest of the methods in each of the $ h $ steps ahead, with $ h \in\{1,2,3\} $. When there is a statistically significant difference between two methods A and B in favor of method A, then method A obtains a `win' while method B obtains a `loss'. In this way, it is possible to perform a win-loss ranking to determine which method is statistically best. Table \ref{tab:ranking} shows the win-loss ranking of the four methods for the Italian city and Lorca air quality problems (the win-loss ranking of both problems coincide). From this table it can be deduced that EFS-LSTM-MOEA is statistically better than the rest of the FS methods, that CancelOut and EAR-FS are statistically better than the conventional LSTM, and that there are no statistically significant differences between CancelOut and EAR-FS.

\begin{table}[!h]
	\centering
	\footnotesize
		\begin{tabular}{cccc}
			\hline
			\textbf{Method}  & \textbf{Win} & \textbf{Loss} & \textbf{Win -- Loss} \\ \hline
			EFS-LSTM-MOEA 	  & 9.0 & 0.0 & 9.0 \\
			CancelOut 	      & 3.0 & 3.0 & 0.0 \\
			EAR-FS 	          & 3.0 & 3.0 & 0.0 \\
			Conventional LSTM & 0.0 & 9.0 & -9.0 \\
			 \hline
		\end{tabular}
	\caption{Win-loss ranking of statistically significant differences with Diebold-Mariano test for the Italian city and Lorca air quality problems.}
	\label{tab:ranking}
\end{table}

\item[--]  EFS-LSTM-MOEA has selected approximately 50\%  of the attributes on average for the two air quality problems and for the 10 runs. CancelOut and EAR-FS have selected a slightly lower percentage of attributes than EFS-LSTM-MOEA, above 40\%. Table \ref{tab:num-attributes} shows the average number of attributes selected and the average percentage of the three FS methods.

\begin{table}[!h]
	\centering
	\footnotesize
		\begin{tabular}{ccc}
			\hline
			\textbf{Method}  & \textbf{\begin{tabular}[c]{c} Average  number\\ of attributes\end{tabular}} & \textbf{\begin{tabular}[c]{c}Average \\ percentage\end{tabular}} \\ \hline
			EFS-LSTM-MOEA 	  & 19.68 & 50.64\%  \\
			CancelOut 	      & 15.90 & 48.18\% \\
			EAR-FS 	          & 14.05 & 43.58\%  
	\\ \hline
		\end{tabular}
		\caption{Average number of attributes selected by the FS methods and average percentage.}
		\label{tab:num-attributes}
	\end{table}

\item[--]  EFS-LSTM-MOEA gives importance of 1 (maximum) to the attributes Lag\_NOxGT\_1, Lag\_PT08.S1(CO)\_2, Lag\_PT08.S2(NMHC)\_1, Lag\_PT08.S1(CO)\_1, Lag\_NOxGT\_2, Lag\_PT08.S3(NOx)\_1 and Lag\_NO2(GT)\_3 for the Italian city air quality problem, and Lag\_RS\_3
Lag\_NO2\_1 and Lag\_NOX\_1 for the Lorca air quality problem. This means that these attributes have been selected in all the Pareto Front models identified by EFS-LSTM-MOEA, and are therefore relevant in all the objective functions $\mathcal{F}_{R_k}$ of the problems, $k=1,\ldots,n$, that is, for the $ n $ partitions into which the data was divided. Figures \ref{fig:fsNOX} and \ref{fig:fsO3} graphically show the importance of the attributes selected with EFS-LSTM-MOEA for the Italian city and Lorca air quality problems respectively.

\begin{figure}[!h]
	\centering
	\resizebox{0.45\textwidth}{!}{\includegraphics[width=\textwidth]{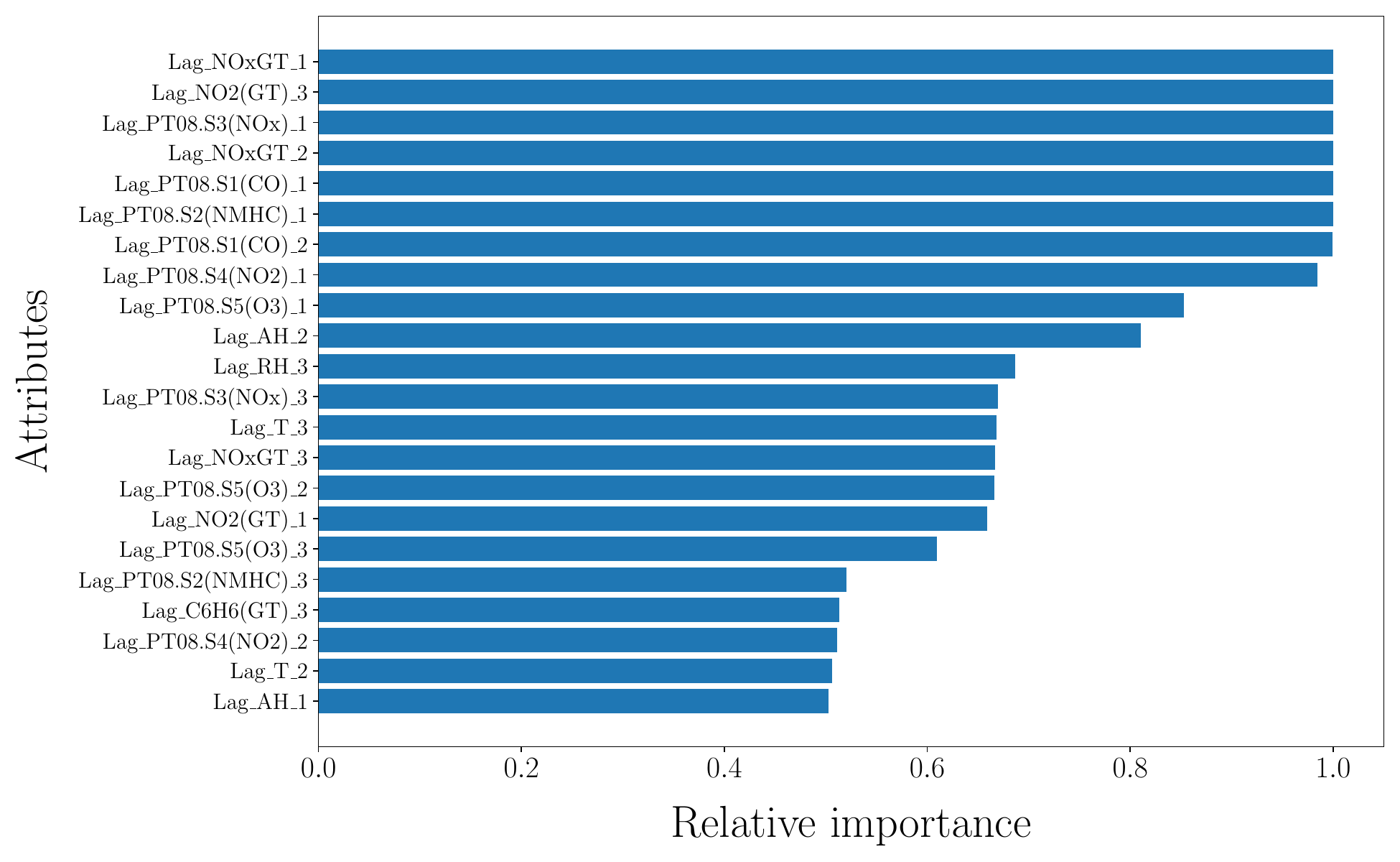}}
	\caption{Feature importance with EFS-LSTM-MOEA for the Italian city air quality problem.}
	\label{fig:fsNOX}
\end{figure}

\begin{figure}[!h]
	\centering
	\resizebox{0.45\textwidth}{!}{\includegraphics[width=\textwidth]{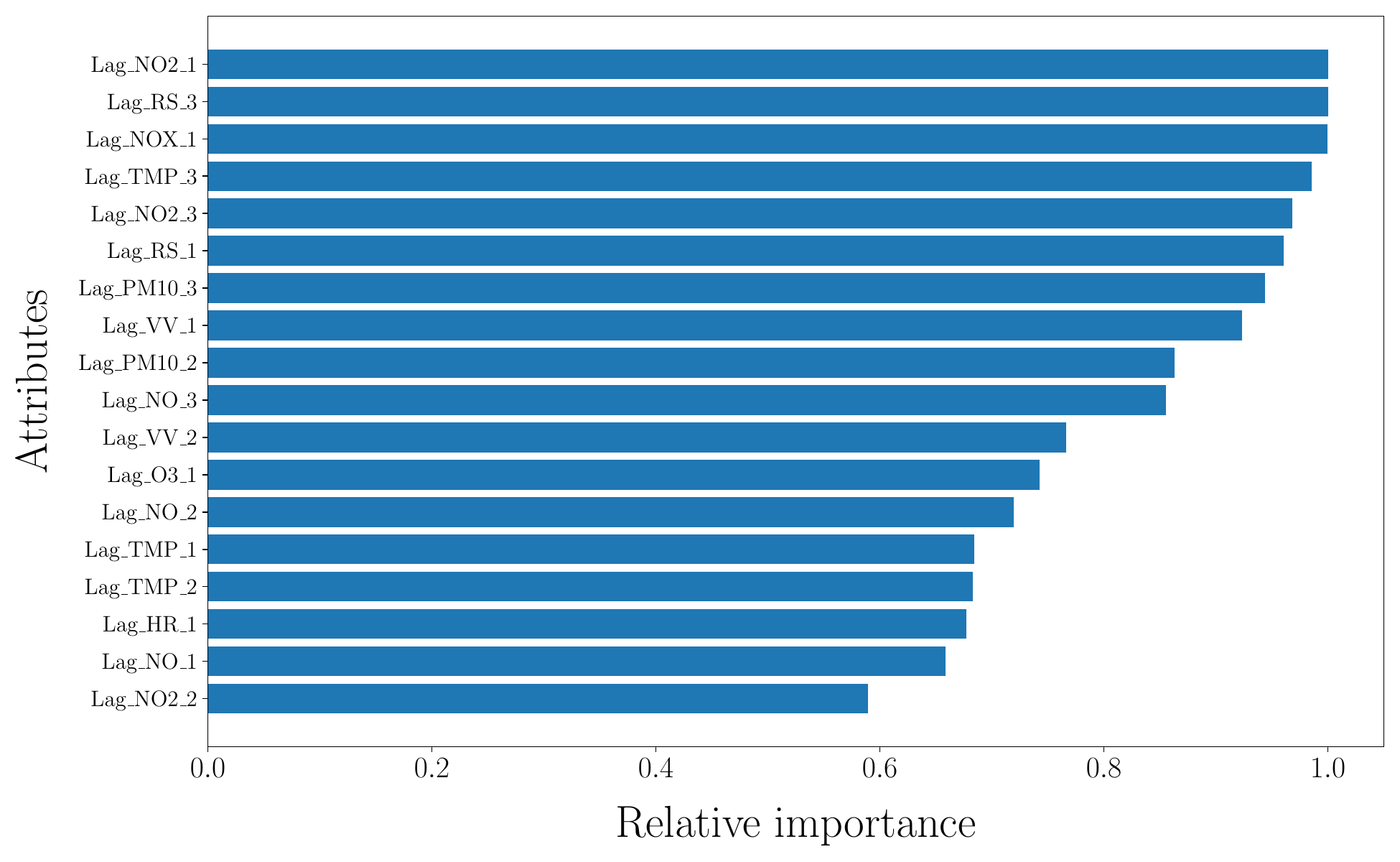}}
	\caption{Feature importance with EFS-LSTM-MOEA for the Lorca air quality problem.}
	\label{fig:fsO3}
\end{figure}

\item[--]  The main drawback of EFS-LSTM-MOEA is the learning time required. Table \ref{tab:runtimes} shows the average run times of the methods analyzed in this paper\footnote{Experiments were run on a computer with motherboard MR91-FS0-00 (R281-3C2-00), memory 12 x 32 GiB DIMM DDR4 2933 MHz (18ASF4G72PDZ-2G9B2), CPU 2 x Intel Xeon Gold 6226R.}. The extended learning time is justified by the significantly improved performance of EFS-LSTM-MOEA compared to conventional LSTM networks, such as better RMSE, reduced overfitting, and improved generalization on analyzed datasets. The longer learning time is a trade-off for the complexity introduced by the evolutionary approach.  The evolutionary process inherently requires more computational resources due to its iterative and search-based nature, but the benefits in terms of predictive ability  outweigh the time cost. The method is well-suited for scenarios where learning can be performed in a batch mode, making it practical for tasks that don't require real-time decision-making. The longer learning time can be considered during model training and optimization phases, which may not be as time-sensitive as real-time inference. Once the model is trained, it can be deployed for fast predictions.  On the other hand, computing systems continue to advance, the inconvenience of longer learning times may be alleviated. With the continuous improvement of hardware and parallel processing capabilities, the efficiency of the evolutionary approach may improve over time.

\begin{table}[!h]
	\centering
	\footnotesize
	\begin{tabular}{cccc}
		\hline
		\textbf{Method}  & \textbf{Italy city} & \textbf{Lorca} & \textbf{Average} \\ \hline
		EFS-LSTM-MOEA & 926.6 & 982.7 & 954.65 \\
		Conventional LSTM & 0.59  & 0.15 & 0.37 \\
		CancelOut & 0.4 & 0.1 & 0.25 \\
		EAR-FS & 1.26 & 0.99 & 1.13 \\ \hline
	\end{tabular}
	\caption{Average runtime (in minutes) of the 10 runs of the methods.}
	\label{tab:runtimes}
\end{table}

\item[--] In relation to the MOEA, the evolution of the \textit{hypervolume} \cite{Zit02} shows an ascending curve from the beginning to the end of the algorithm, which is indicative of good behaviour of EFS-LSTM-MOEA. The hypervolume metric measures both convergence and diversity of a population of non-dominated solutions. Figures \ref{fig:AQevol} and \ref{fig:Lorcaevol} correspond to the hypervolume evolution of the run that has produced the best result for the Italian city and Lorca air quality problems respectively. Figures \ref{fig:AQpareto} and \ref{fig:Lorcapareto} show the \textit{parallel coordinate chart} of the Pareto front obtained with EFS-LSTM-MOEA in the run that has produced the best result for the Italian city and Lorca air quality problems respectively. A parallel coordinate chart is a visualization technique that can be used to represent multi-dimensional data, making it suitable for visualizing the Pareto front obtained from an evolutionary algorithm with more than three objective functions. The lines in the charts cover a wide range of values across different axes indicating a diverse set of solutions. Additionally, the lines cross each other frequently, indicating that the algorithm is exploring diverse regions of the search space.

\begin{figure}[!h]
	\centering
	\resizebox{0.45\textwidth}{!}{\includegraphics[width=\textwidth]{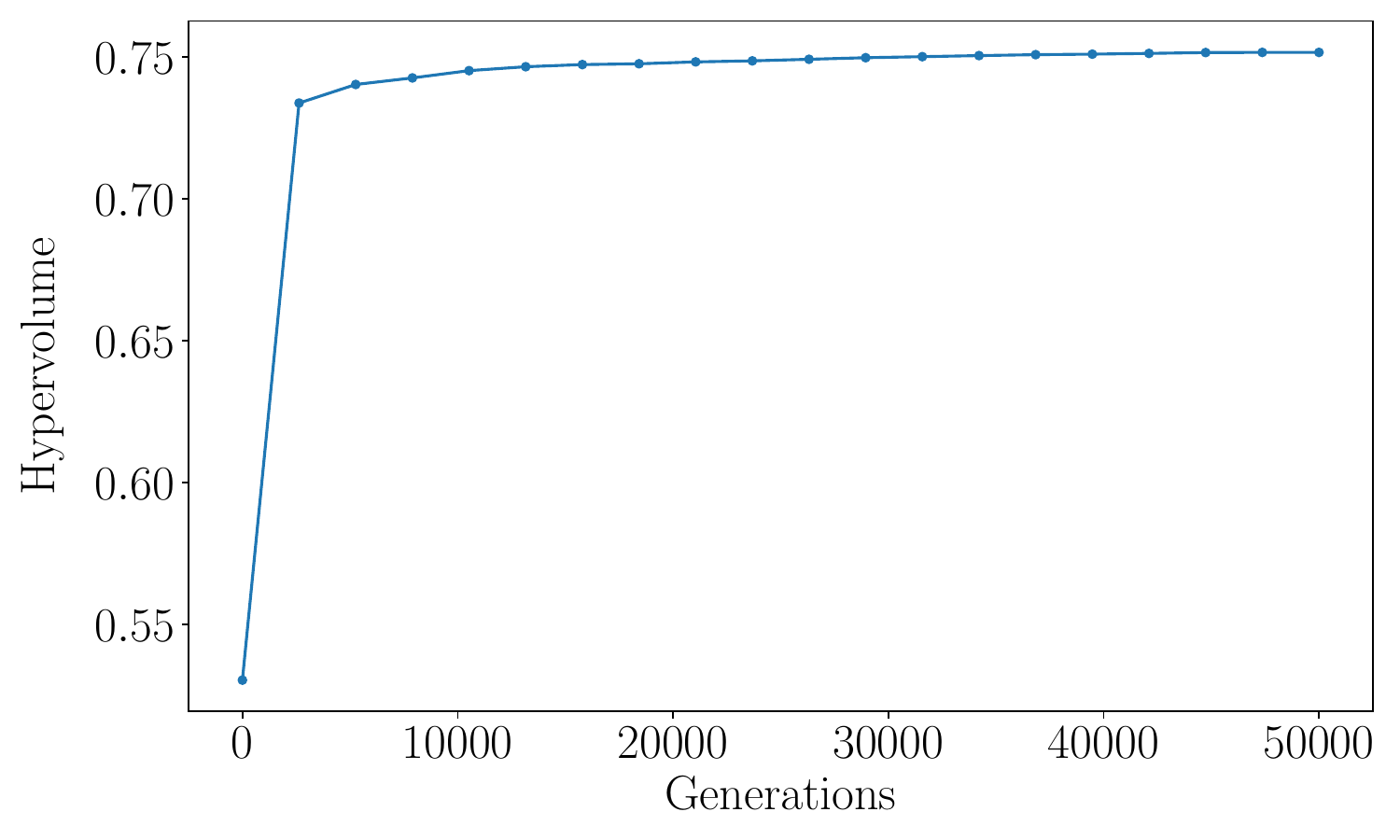}}
	\caption{Hypervolume evolution using EFS-LSTM-MOEA for the Italian city air quality problem.}
	\label{fig:AQevol}
\end{figure}

\begin{figure}[!h]
	\centering
	\resizebox{0.45\textwidth}{!}{\includegraphics[width=\textwidth]{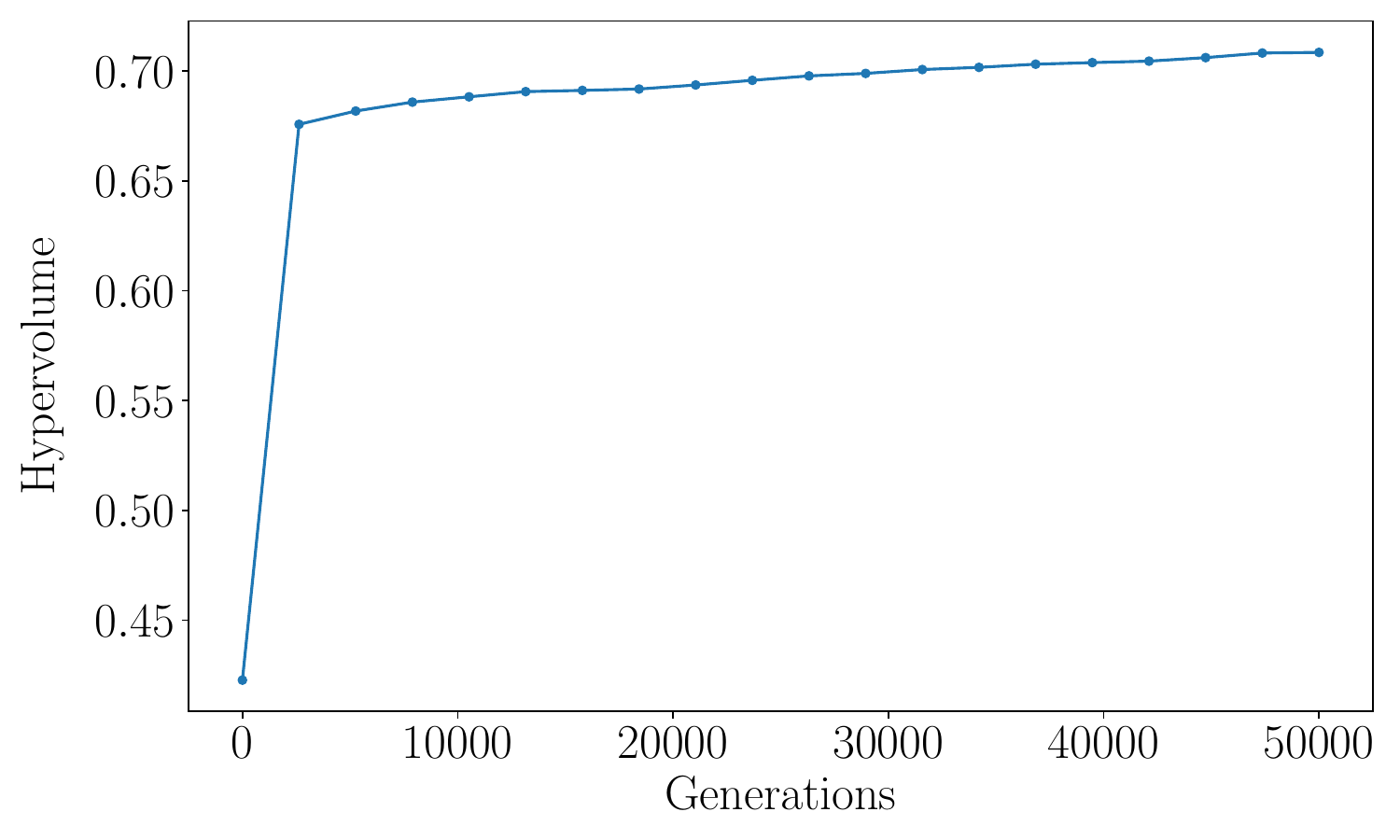}}
	\caption{Hypervolume evolution using EFS-LSTM-MOEA for the Lorca air quality problem.}
	\label{fig:Lorcaevol}
\end{figure}

\begin{figure}[!h]
	\centering
	\resizebox{0.45\textwidth}{!}{\includegraphics[width=\textwidth]{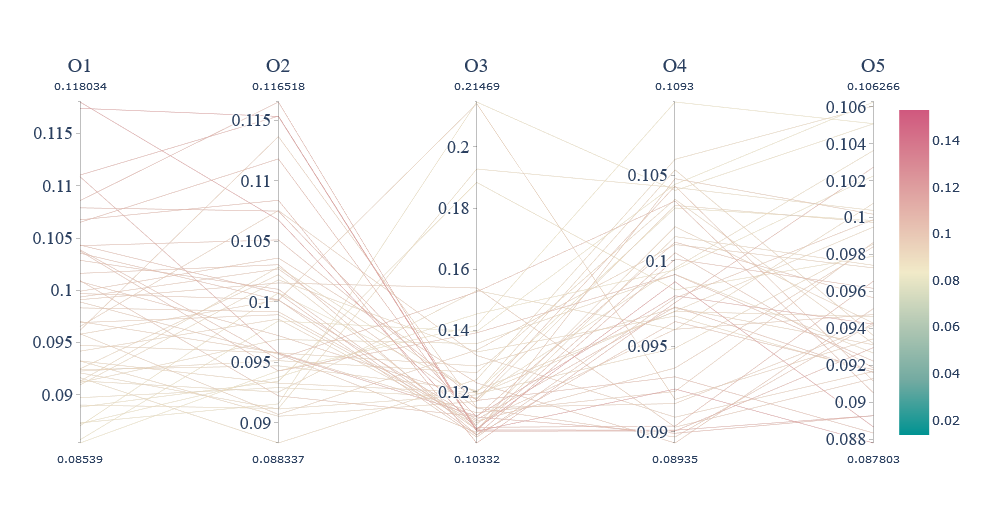}}
	\caption{Parallel coordinates plot of the Pareto front obtained with EFS-LSTM-MOEA for the Italian city air quality problem.}
	\label{fig:AQpareto}
\end{figure}

\begin{figure}[!h]
	\centering
	\resizebox{0.45\textwidth}{!}{\includegraphics[width=\textwidth]{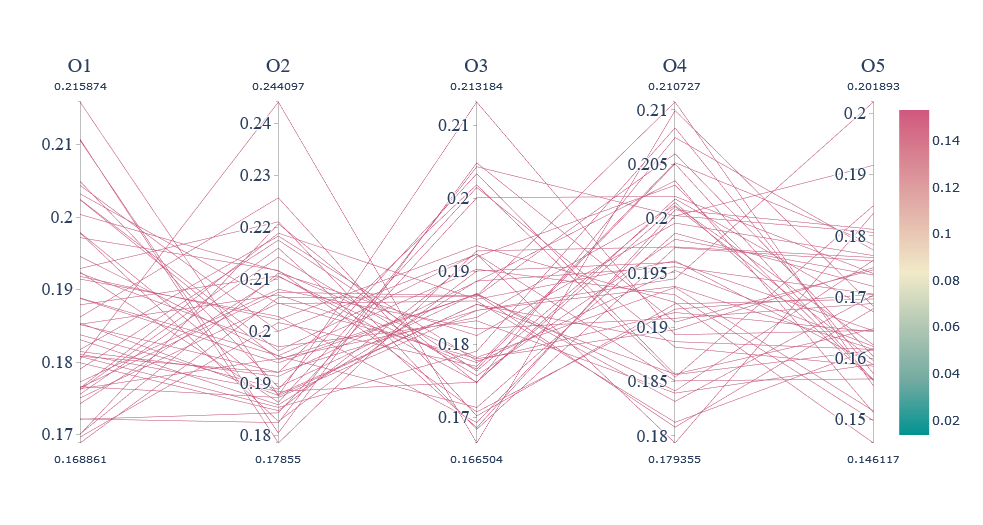}}
	\caption{Parallel coordinates plot of the Pareto front obtained with EFS-LSTM-MOEA for the Lorca air quality problem.}
	\label{fig:Lorcapareto}
\end{figure}

\item[--] Regarding the forecast model built with EFS-LSTM-MOEA we can highlight, in addition to its high generalization capacity already mentioned, its robustness in multi-step ahead predictions. Although the recursive strategy for multi-step ahead prediction tends to accumulate prediction errors between successive steps ahead, the predictions made by the model over the entire future prediction horizon in both air quality problems remain stable without significant losses, as shown in Tables \ref{tab:hstepsNOX} and \ref{tab:hstepsNO2}. To close this section of analysis of results we have included in Appendix A (Figures \ref{fig:predictionmodeltrainO3} to \ref{fig:predictionmodeltestO3}) the predictions on training dataset and test dataset for 1, 2 and 3 steps ahead of the models obtained with EFS-LSTM-MOEA in the two air quality problems analyzed. 

\begin{table}[!h]
	\centering
	\footnotesize
		\begin{tabular}{ccccc}
			\hline
			\textbf{\begin{tabular}[c]{@{}c@{}}Evaluation\\ dataset\end{tabular}} & \textbf{\begin{tabular}[c]{@{}c@{}}Performance\\ metric\end{tabular}} & \textbf{\begin{tabular}[c]{@{}c@{}}1-step\\ ahead\end{tabular}} & \textbf{\begin{tabular}[c]{@{}c@{}}2-steps\\ ahead\end{tabular}} & \textbf{\begin{tabular}[c]{@{}c@{}}3-steps\\ ahead\end{tabular}} \\ \hline
			\multirow{2}{*}{$R$} & RMSE   &  0.10336 &  0.10493 &  0.10726   	 	 	\\
			 	& MAE    &  0.07439 &  0.07643 &  0.07783   	 		\\
			\hline
			\multirow{2}{*}{$T$} & RMSE   &  0.09808 &  0.09723 &  0.09987     	 	\\
			 	& MAE    &  0.06706 &  0.06724 &  0.07029   		 	\\
			\hline
		\end{tabular}
	\caption{Results of the best forecast model for the Italian city air quality problem, evaluated on the training dataset $R$ and the test dataset $T$.}
	\label{tab:hstepsNOX}
	
\end{table}

\begin{table}[!h]
	\centering
	\footnotesize
		\begin{tabular}{ccccc}
			\hline
			\textbf{\begin{tabular}[c]{@{}c@{}}Evaluation\\ dataset\end{tabular}} & \textbf{\begin{tabular}[c]{@{}c@{}}Performance\\ metric\end{tabular}} & \textbf{\begin{tabular}[c]{@{}c@{}}1-step\\ ahead\end{tabular}} & \textbf{\begin{tabular}[c]{@{}c@{}}2-steps\\ ahead\end{tabular}} & \textbf{\begin{tabular}[c]{@{}c@{}}3-steps\\ ahead\end{tabular}} \\ \hline
			\multirow{2}{*}{$R$} & RMSE   &  0.18772 &  0.19003 &  0.18542   	 	 	\\
			 	& MAE    &  0.15405 &  0.15551 &  0.15144   	 		\\
			\hline
			\multirow{2}{*}{$T$} & RMSE   &  0.17346 &  0.17326 &  0.17111     	 	\\
			  	& MAE    &  0.13713 &  0.13597 &  0.13485   		 	\\
			\hline
		\end{tabular}
	\caption{Results of the best forecast model for the Lorca air quality problem, evaluated on the training dataset $R$ and the test dataset $T$.}
	\label{tab:hstepsNO2}
	
\end{table}

\end{itemize}

\section{Conclusions and future work}  
\label{Conclusion}
In this study, we introduced a novel approach for embedded feature selection in LSTM networks using a multi-objective evolutionary algorithm. Our method addressed the challenges of overfitting and enhanced the generalization ability of conventional LSTMs, demonstrating superior performance compared to state-of-the-art CancelOut and EAR-FS methods. By optimizing the weights and biases of the LSTM through a partitioned multi-objective evolutionary algorithm, we identified a set of non-dominated forecast models that were subsequently employed in a stacking-based ensemble learning framework to construct a robust meta-model. The frequency of selection of each attribute in the set of non-dominated forecast models allows us to calculate its importance

The key findings of our experiments, conducted on two distinct air quality time series datasets from Italy and the southeast of Spain, showcased the efficacy of our proposed method in achieving improved forecasting accuracy and improving model interpretability. The reduction in overfitting, coupled with outperformance against existing techniques, highlights the potential of our approach for enhancing the reliability, precision and interpretability of time series forecasting tasks. 

As we look towards future work, several avenues emerge for the further development and expansion of our proposed technique. One promising direction involves extending our methodology to transformer neural networks, which have gained significant attention in recent years due to their superior performance in various natural language processing and time series tasks. The inherent parallelization capabilities and attention mechanisms of transformers make them an intriguing platform for the integration of our feature selection technique.
Additionally, exploring the scalability and adaptability of our method to diverse domains and datasets remains an essential aspect of future research. Investigating the generalization of our approach to real-world applications beyond air quality forecasting could provide valuable insights into the versatility and robustness of the proposed feature selection mechanism.
Furthermore, refining the multi-objective evolutionary algorithm parameters and exploring alternative optimization techniques could contribute to the fine-tuning and optimization of our approach. Experimenting with different partitioning strategies and evaluating their impact on the model's performance could provide a deeper understanding of the interplay between data partitioning and feature selection in LSTM networks.

In conclusion, our proposed feature selection method embedded in LSTM, optimized by a multi-objective evolutionary algorithm, demonstrates promising results and opens avenues for further advancements in time series forecasting. The extension of this technique to transformer neural networks and the exploration of its applicability across diverse domains present exciting prospects for future research, contributing to the ongoing evolution of intelligent forecasting models.

\section*{Acknowledgements}
This paper is funded by the CALM-COVID19 project (Ref:
PID2022-136306OB-I00), grant funded by Spanish Ministry of
Science and Innovation and the Spanish Agency for Research.






\bibliographystyle{elsarticle-num}
\bibliography{references}

\newpage
\clearpage
\appendix

\section{Abbreviations}

\begin{table}[!h]
	\centering
	\label{tab:abbreviations}
	\resizebox{0.87\textwidth}{!}{
		\begin{tabular}{ll}
			\hline
			\textbf{Abbreviation} & \textbf{Meaning}                                                   \\ 
			\hline
			AFS					  & Attention-based Feature Selection \\
			AH					  & Absolute Humidity \\
			$C_6H_6$			  & Benzene															   \\
			$C_7H_8$			  & Toluene															   \\
			CNN                   & Convolutional Neural Network                                       \\
			CO					  & Carbon Monoxide \\
			D-AFS				  & Dual-world embedded Attentive FS \\
			DD   				  & Wind Direction \\
			EAR-FS & External Attention-Based Feature Ranker for Large-Scale Feature Selection  \\
			EFS-LSTM-MOEA		  & Embedded Feature Selection in LSTM networks with Multi-Objective Evolutionary Algorithms \\
			FS					  & Feature Selection \\
			HR 					  & Relative Humidity \\
			IFSE-NEAT			  & Incremental Feature Selection Embedded in NEAT \\	
			IL 					  & Incremental Learning \\
			LSTM                  & Long Short-Term Memory                                             \\
			MAE					  & Mean Absolute Error												   \\
			MLP 				  & Multi-layer Perceptron \\
			MOEA                  & Multi-Objective Evolutionary Algorithm                             \\
			NEAT				  & Neuroevolution of Augmenting Topologies \\
			NFS					  & Neural Feature Selection \\
			NMHC				  & Non Metanic HydroCarbons \\
			$NO$ 				  & Nitrogen Monoxide \\
			$NO_2$                & Nitrogen Dioxide                              \\
			$NO_X$				  & Nitrogen Oxides	\\
			NSGA-II               & Non-dominated Sorting Genetic Algorithm II                         \\
			$O_3$				  & Ozone															   \\
			$PM_{10}$             & Particulate Matter (with an aerodynamic diameter smaller than) 10     \\
			PRB 				  & Atmospheric Pressure \\
			RF                    & Random Forest                                                      \\
			RMSE                  & Root Mean Squared Error                                            \\
			RNN 				  & Recurrent Neural Network \\
			RS 					  & Solar Radiation \\
			SBS					  & Sensitivity Based Selection \\
			$SO_2$				  & Sulfur dioxide													   \\
			SVM                   & Support Vector Machine                                             \\
			SWPA				  & Stepwise Weight Pruning Algorithm \\
			TMP 				  & Temperature \\
			TSFS				  & Teacher-Student Feature Selection \\
			UCI 				  & University of California Irvine \\
			VV 					  & Wind Speed \\
			XIL 				  & Xylene \\
			\hline
		\end{tabular}
	}
\end{table}

\newpage
\clearpage

\section{Predictions of forecasting models}
\label{apped:performancepred}

\vskip 2.56cm
\begin{figure}[h]
	\centering
	
	\begin{subfigure}[b]{0.47\textwidth}
		\centering
		\includegraphics[width=1\textwidth]{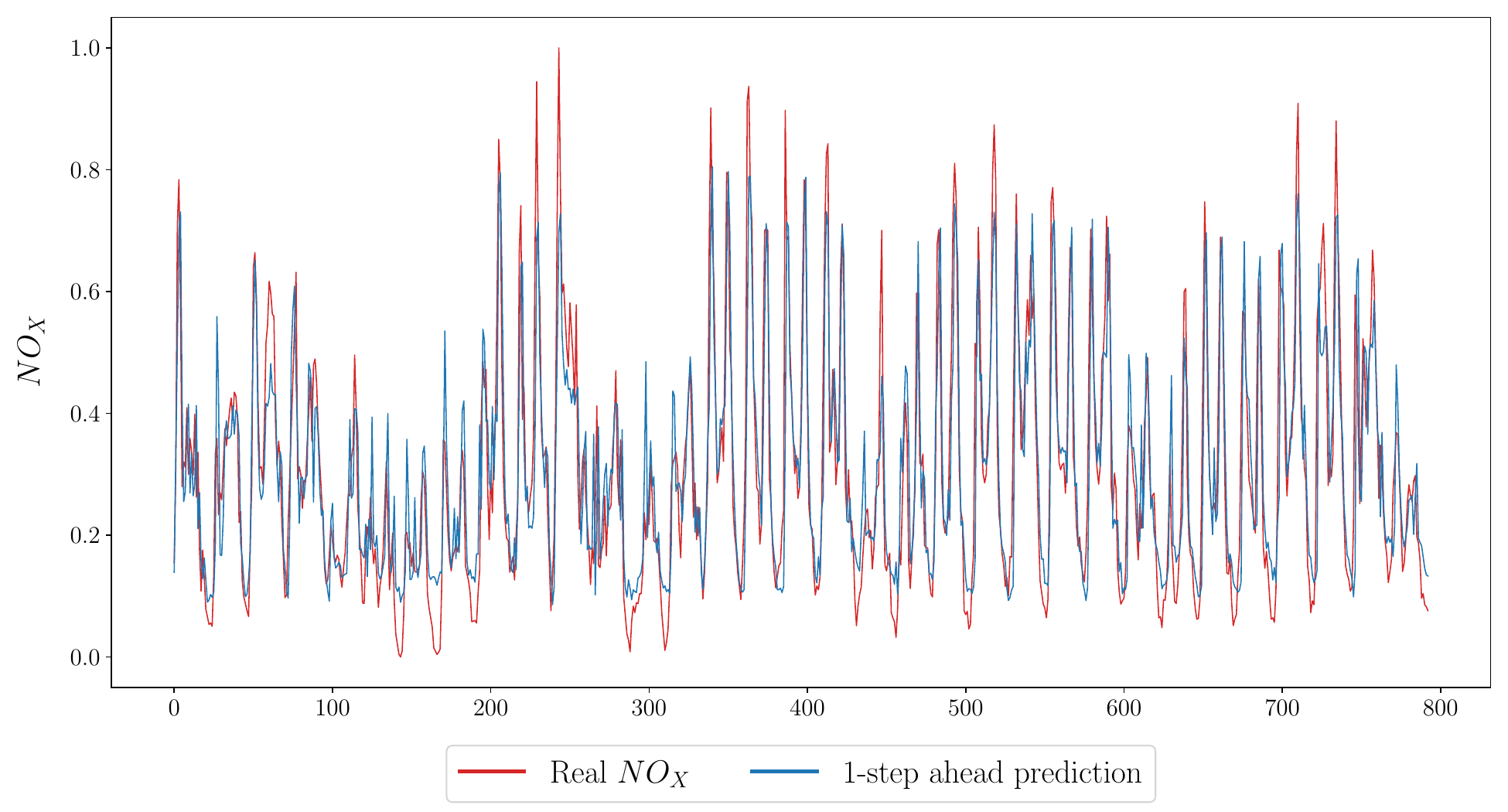}
		\caption{1-step ahead}
	\end{subfigure}
	\begin{subfigure}[b]{0.47\textwidth}
		\centering
		\includegraphics[width=1\textwidth]{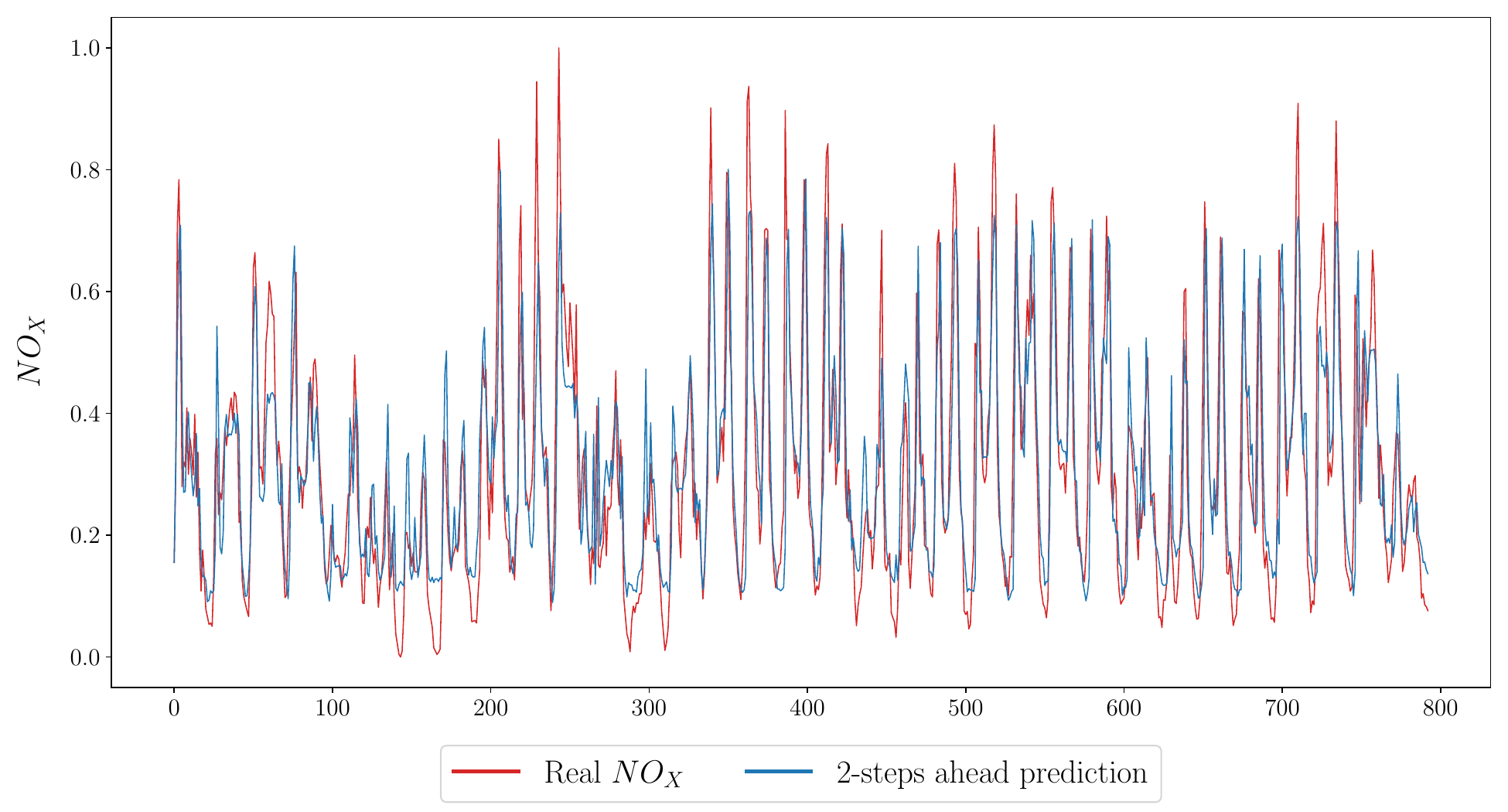}
		\caption{2-steps ahead}
	\end{subfigure}
	
	\begin{subfigure}[b]{0.47\textwidth}
		\centering
		\includegraphics[width=1\textwidth]{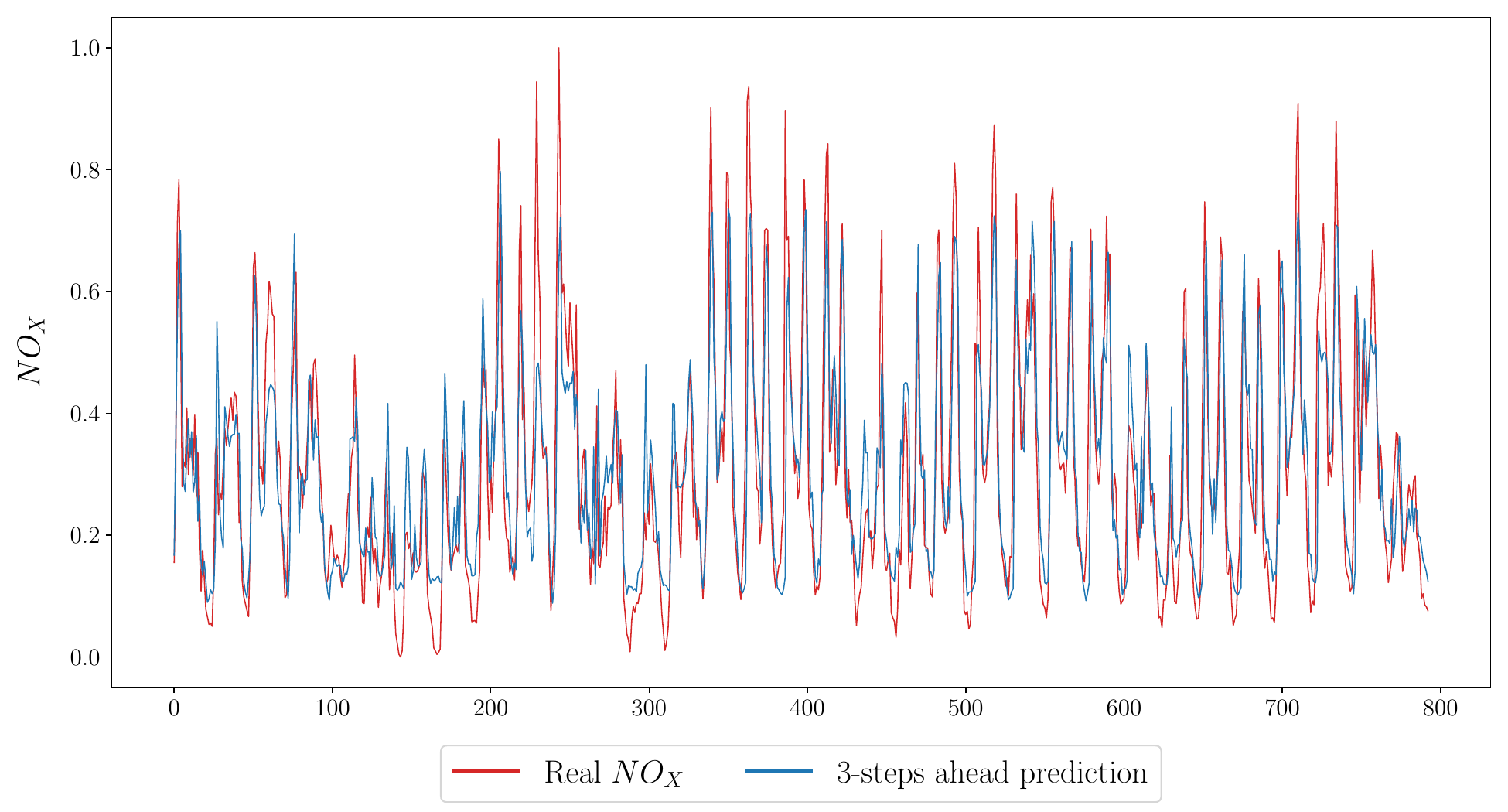}
		\caption{3-steps ahead}
	\end{subfigure}
	
	\caption{1, 2 and 3 steps ahead predictions of the forecast model obtained with EFS-LSTM-MOEA on training dataset for the Italian city air quality problem.}
\end{figure}

\begin{figure}[h]
	\centering
	
	\begin{subfigure}[b]{0.47\textwidth}
		\centering
		\includegraphics[width=1\textwidth]{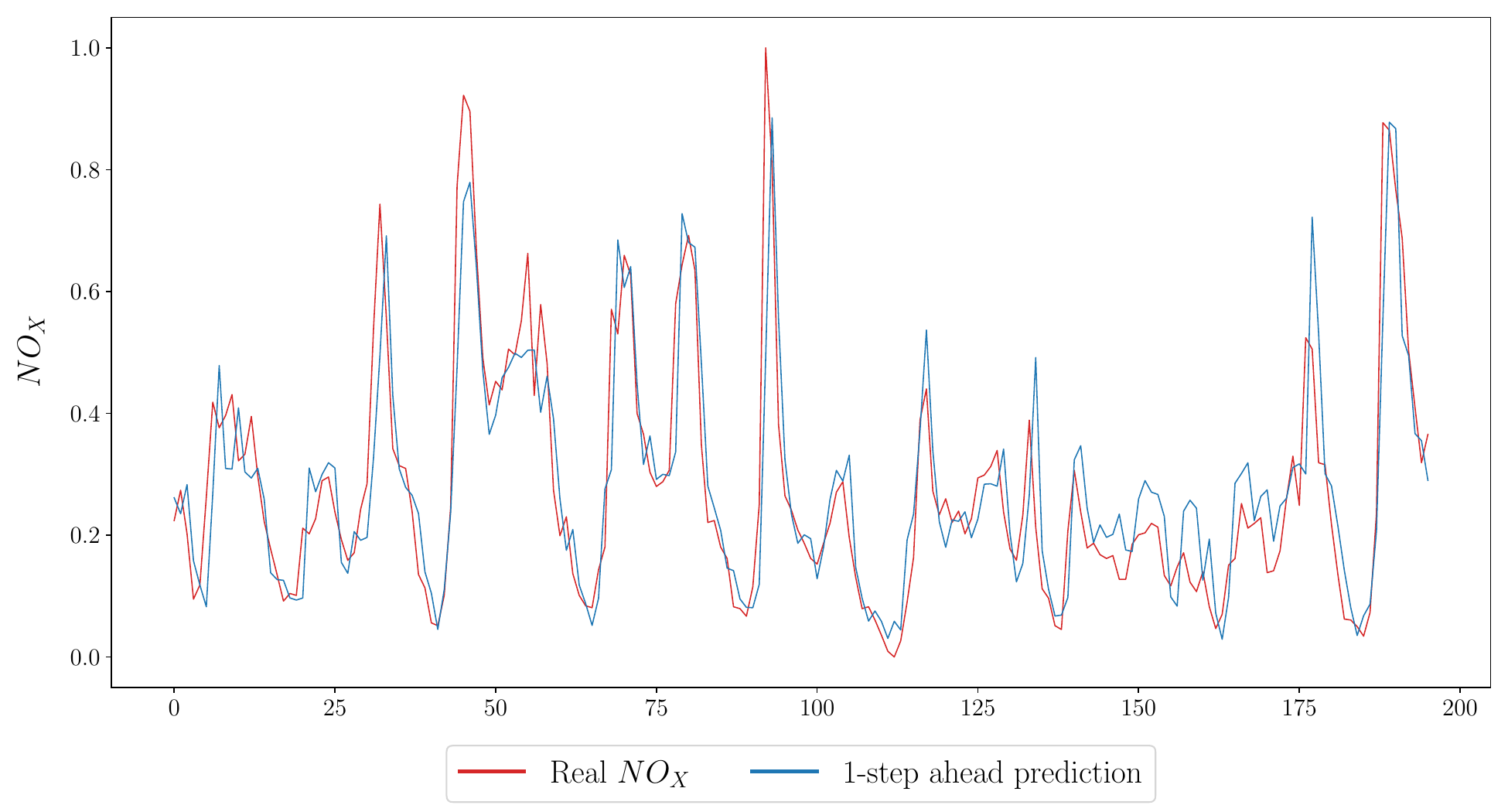}
		\caption{1-step ahead}
	\end{subfigure}
	\begin{subfigure}[b]{0.47\textwidth}
		\centering
		\includegraphics[width=1\textwidth]{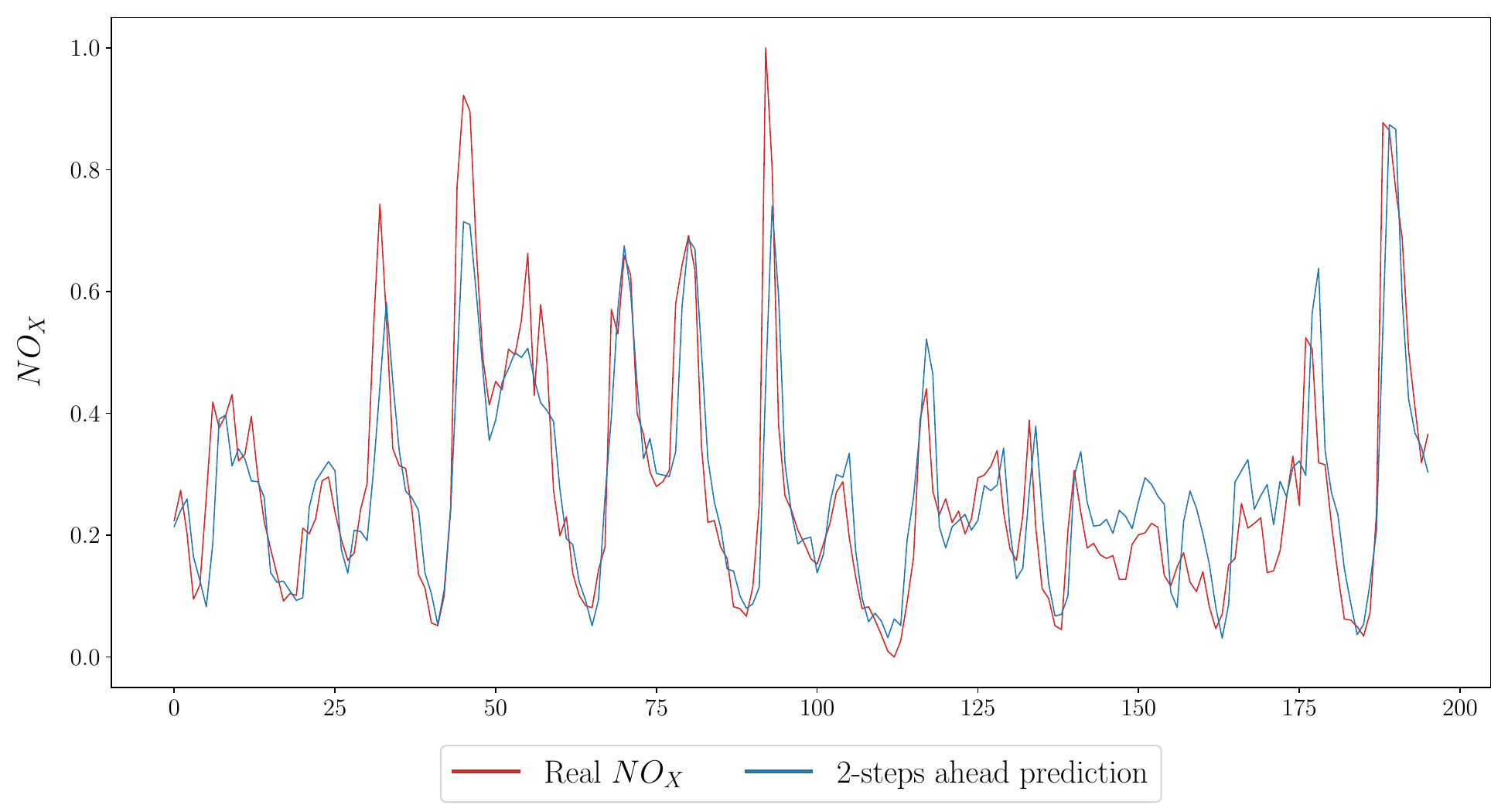}
		\caption{2-steps ahead}
	\end{subfigure}
	
	\begin{subfigure}[b]{0.47\textwidth}
		\centering
		\includegraphics[width=1\textwidth]{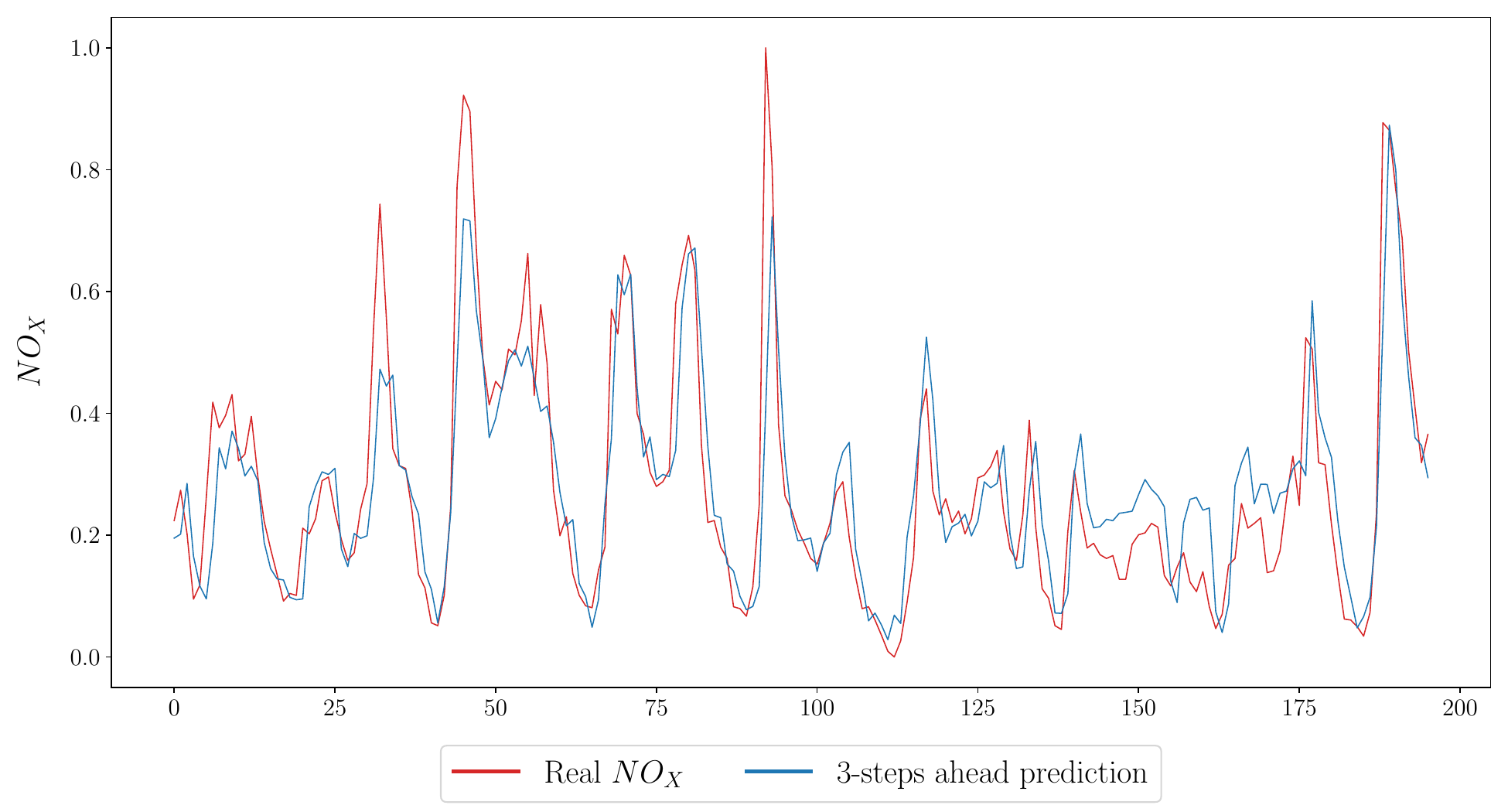}
		\caption{3-steps ahead}
	\end{subfigure}
	
	\caption{1, 2 and 3 steps ahead predictions of the forecast model obtained with EFS-LSTM-MOEA on test dataset for the Italian city air quality problem.}
	\label{fig:predictionmodeltestNOX}
\end{figure}




\begin{figure}[!htpb]
	\centering
	
	\begin{subfigure}[b]{0.47\textwidth}
		\centering
		\includegraphics[width=1\textwidth]{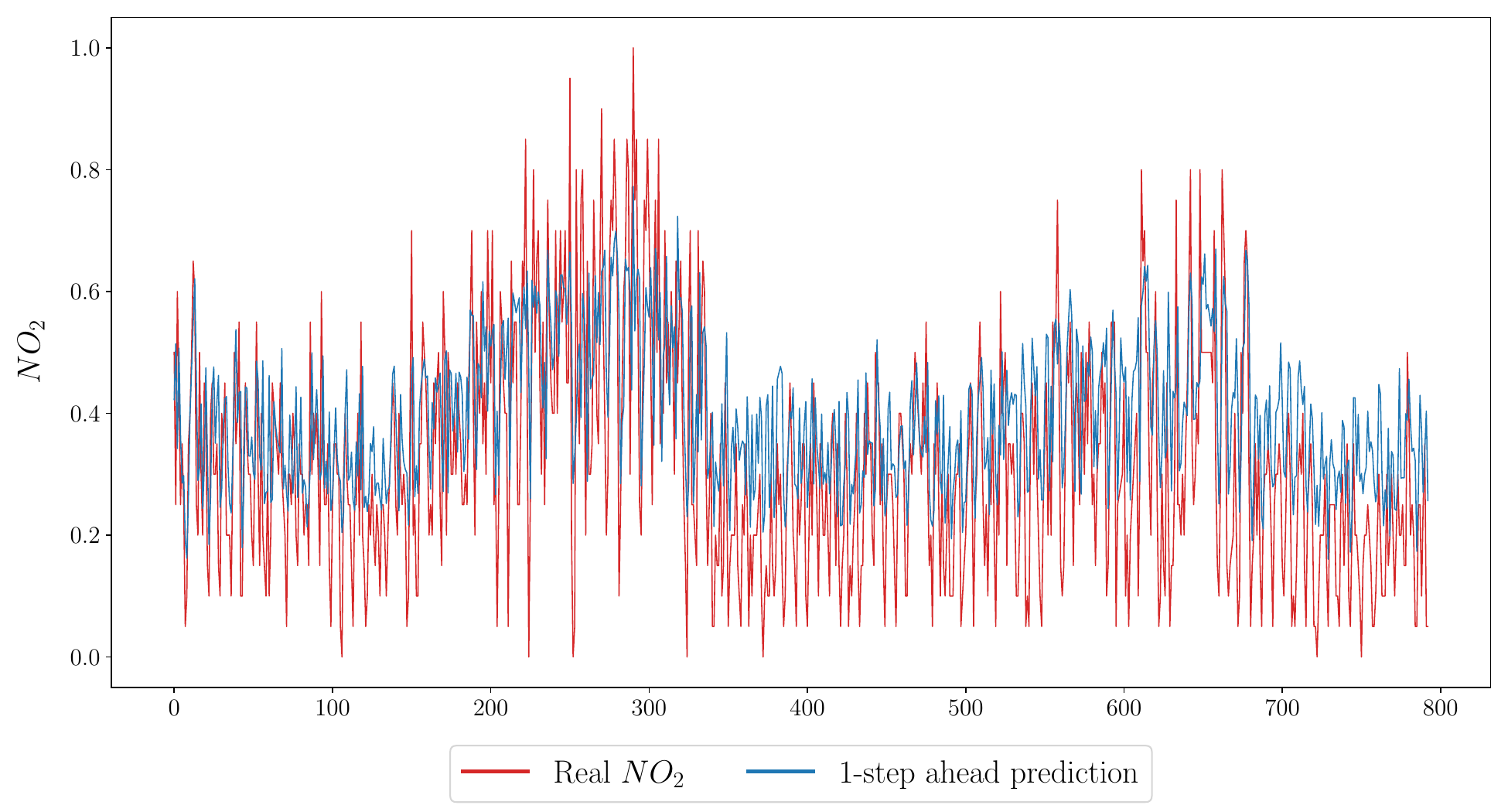}
		\caption{1-step ahead}
	\end{subfigure}
	\begin{subfigure}[b]{0.47\textwidth}
		\centering
		\includegraphics[width=1\textwidth]{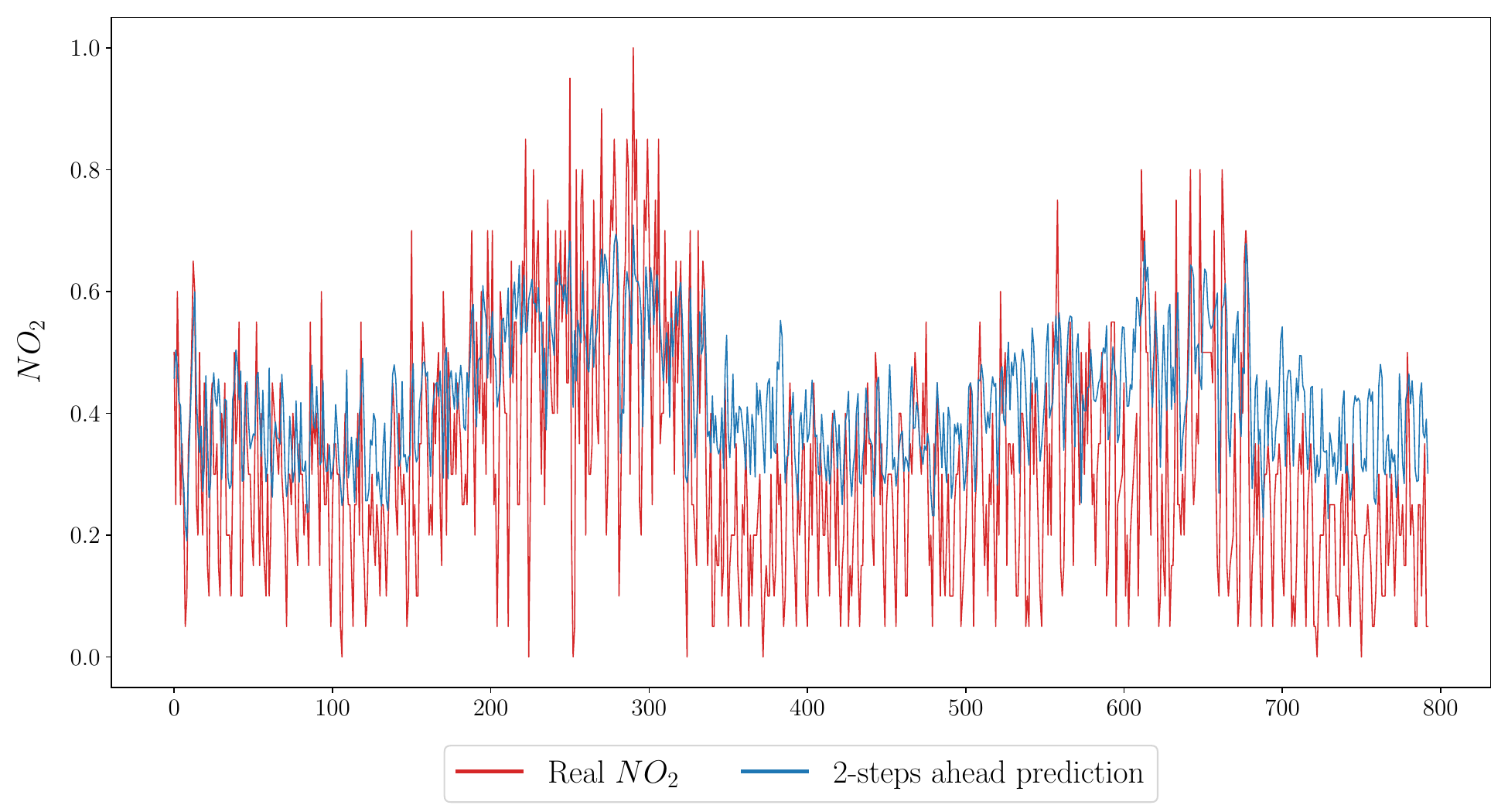}
		\caption{2-steps ahead}
	\end{subfigure}
	
	\begin{subfigure}[b]{0.47\textwidth}
		\centering
		\includegraphics[width=1\textwidth]{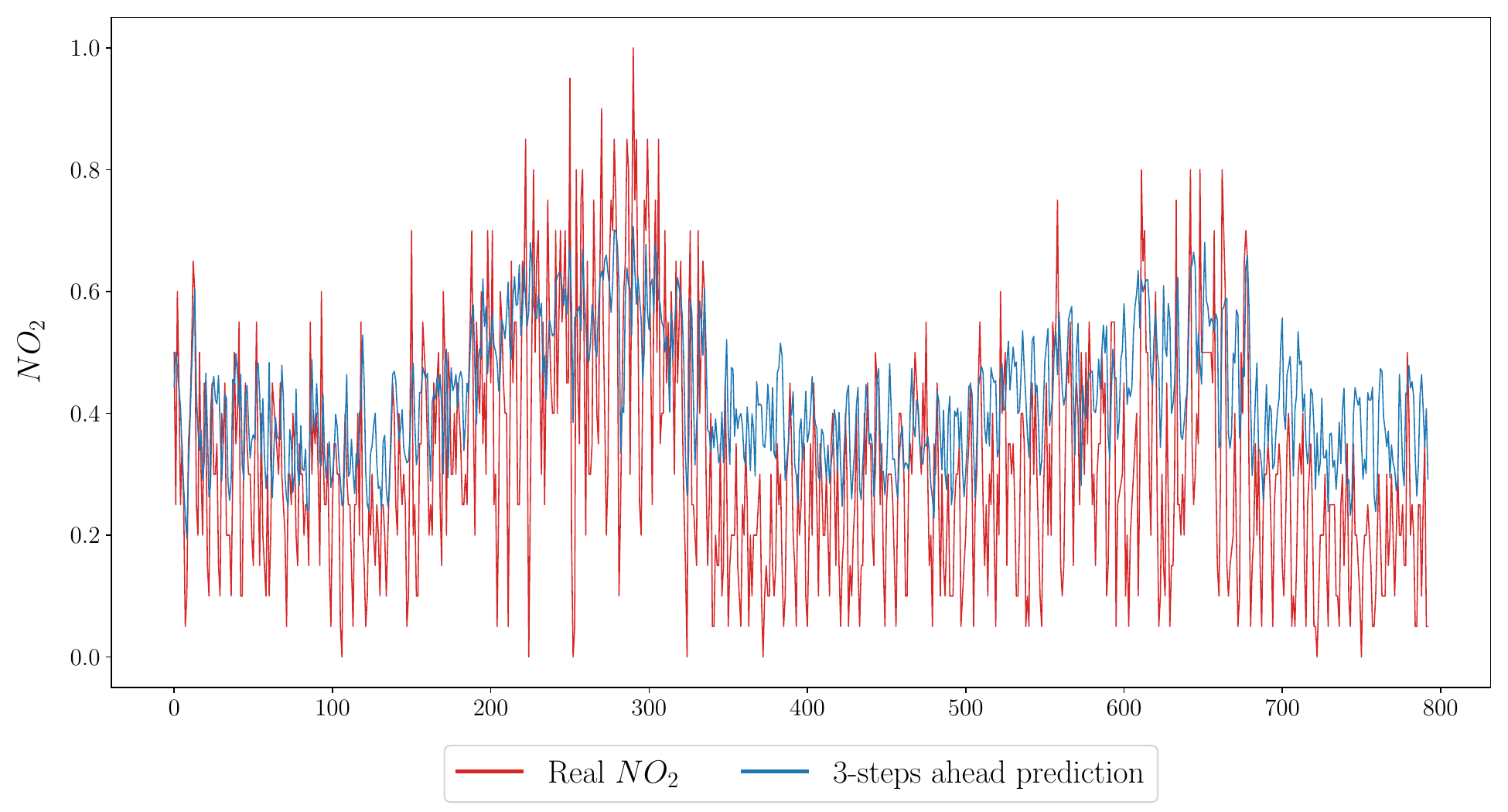}
		\caption{3-steps ahead}
	\end{subfigure}
	
	\caption{1, 2 and 3 steps ahead predictions of the forecast model obtained with EFS-LSTM-MOEA on training dataset for the Lorca air quality problem.}
	\label{fig:predictionmodeltrainO3}
\end{figure}

\begin{figure}[!htpb]
	\centering
	
	\begin{subfigure}[b]{0.47\textwidth}
		\centering
		\includegraphics[width=1\textwidth]{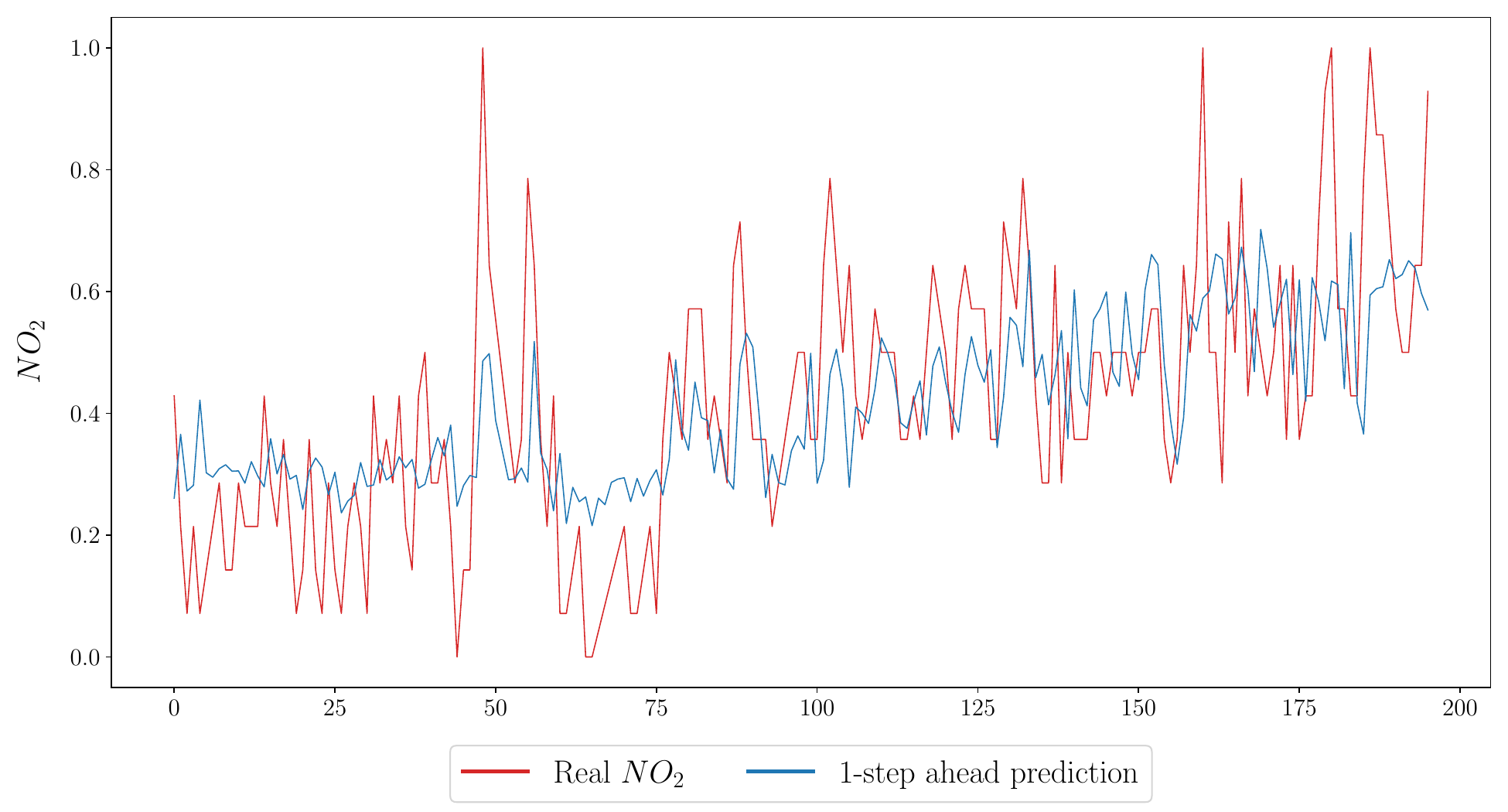}
		\caption{1-step ahead}
	\end{subfigure}
	\begin{subfigure}[b]{0.47\textwidth}
		\centering
		\includegraphics[width=1\textwidth]{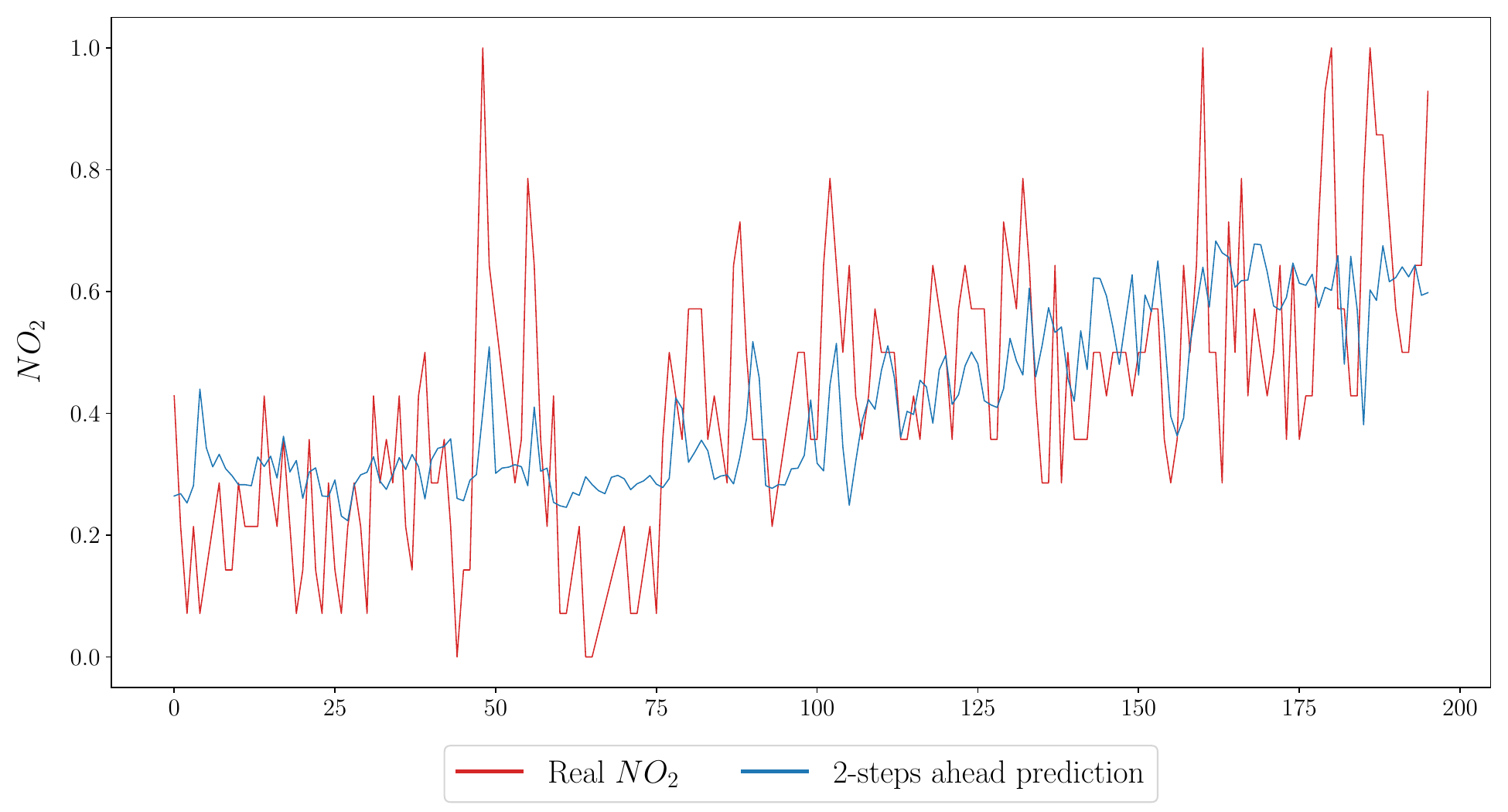}
		\caption{2-steps ahead}
	\end{subfigure}
	
	\begin{subfigure}[b]{0.47\textwidth}
		\centering
		\includegraphics[width=1\textwidth]{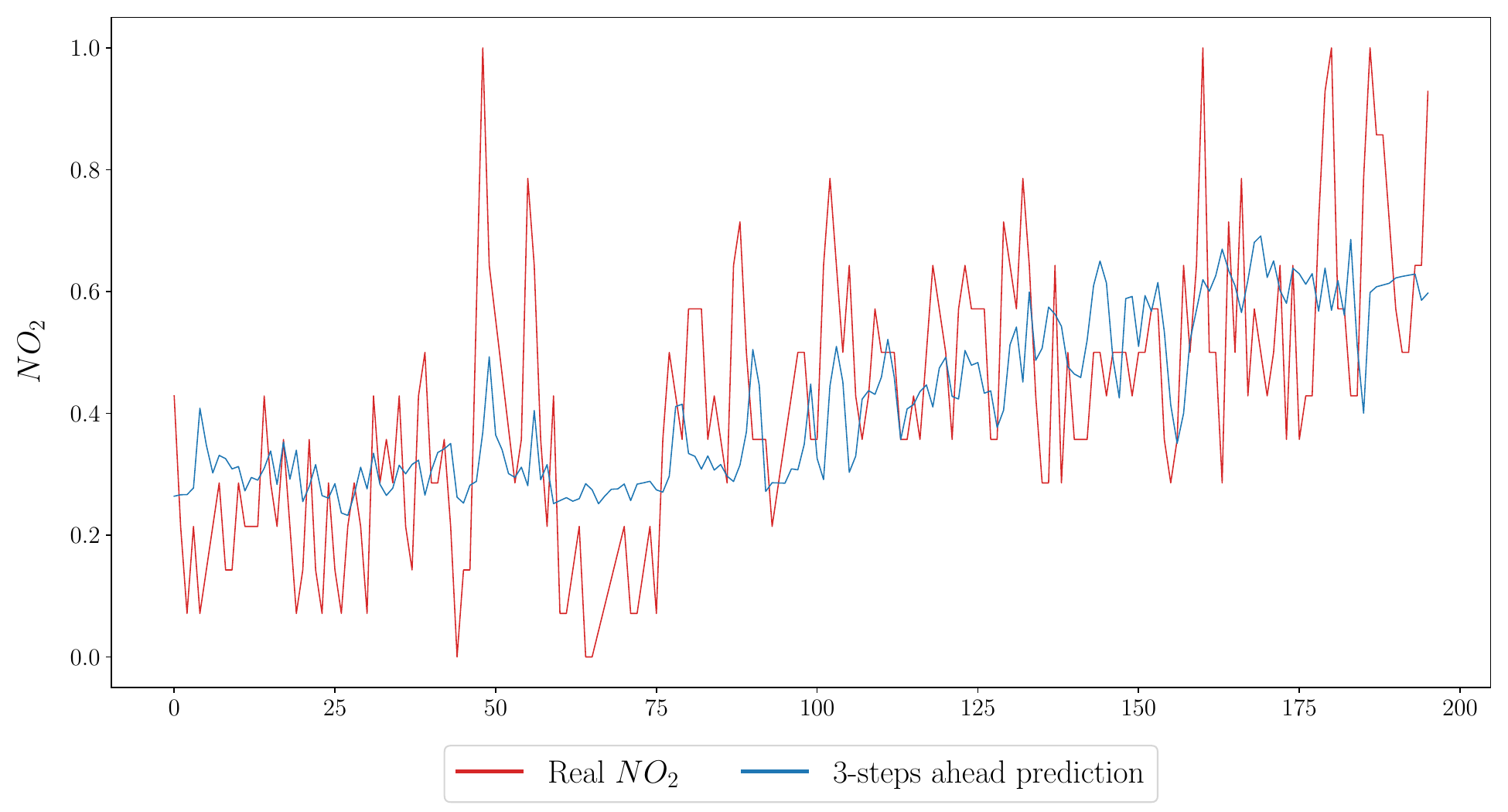}
		\caption{3-steps ahead}
	\end{subfigure}
	
	\caption{T1, 2 and 3 steps ahead predictions of the forecast model obtained with EFS-LSTM-MOEA on test dataset for the Lorca air quality problem.}
	\label{fig:predictionmodeltestO3}
\end{figure}

%

\end{document}